\documentclass[sigconf]{acmart}
\copyrightyear{2023}
\acmYear{2023}
\setcopyright{acmcopyright}\acmConference[KDD '23]{Proceedings of the 29th ACM SIGKDD Conference on Knowledge Discovery and Data Mining}{August 6--10, 2023}{Long Beach, CA, USA}
\acmBooktitle{Proceedings of the 29th ACM SIGKDD Conference on Knowledge Discovery and Data Mining (KDD '23), August 6--10, 2023, Long Beach, CA, USA}
\usepackage[utf8]{inputenc} % allow utf-8 input
\usepackage[T1]{fontenc}    % use 8-bit T1 fonts
\usepackage{hyperref}       % hyperlinks
\usepackage{url}            % simple URL typesetting
\usepackage{booktabs}       % professional-quality tables
\usepackage{amsfonts}       % blackboard math symbols
\usepackage{nicefrac}       % compact symbols for 1/2, etc.
\usepackage{microtype}      % microtypography
\usepackage{float}
\usepackage{natbib}
\usepackage{soul}
\usepackage{graphics}
\usepackage{graphicx}
\usepackage{subfigure}

\usepackage{url}
\usepackage{balance}

\usepackage[ruled,vlined,linesnumbered]{algorithm2e}
\usepackage{enumitem}
\usepackage{paralist}
\usepackage{multirow,multicol,xspace}
\usepackage{amsthm,amsmath}
\usepackage{bbm}
\setitemize{noitemsep,topsep=3pt,parsep=3pt,partopsep=3pt}

\newtheorem{theorem}{Theorem}[section]

\RequirePackage{algorithmic}

\usepackage{cleveref}
\crefname{equation}{Eq.}{Eqs.}
\crefname{table}{Table}{Tables}
\crefname{figure}{Figure}{Figures}
\crefname{section}{Section}{Sections}
\crefname{algorithm}{Algorithm}{Algorithms}

\usepackage{pifont}% http://ctan.org/pkg/pifont
\newcommand{\cmark}{\ding{51}}%
\newcommand{\xmark}{\ding{55}}%

\AtBeginDocument{%
  \providecommand\BibTeX{{%
    \normalfont B\kern-0.5em{\scshape i\kern-0.25em b}\kern-0.8em\TeX}}}

\begin{document}
\title{Semi-Supervised Graph Imbalanced Regression}
\author{Gang Liu} 
\affiliation{%
  \institution{{University of Notre Dame \country{USA}}}
}
\email{gliu7@nd.edu}

\author{Tong Zhao} 
\affiliation{
    \institution{{Snap Inc. \country{USA}}}
}
\email{tzhao@snap.com}

\author{Eric Inae} 
\affiliation{%
    \institution{{University of Notre Dame \country{USA}}}
}
\email{einae@nd.edu}

\author{Tengfei Luo} 
\affiliation{%
    \institution{{University of Notre Dame \country{USA}}}
}
\email{tluo@nd.edu}

\author{Meng Jiang} 
\affiliation{%
    \institution{{University of Notre Dame \country{USA}}}
}
\email{mjiang2@nd.edu}

\newcommand{\method}{\textsc{SGIR}\xspace}

% methods SGIR
\newcommand{\darp}{\textsc{DARP}\xspace}
\newcommand{\daso}{\textsc{DASO}\xspace}
\newcommand{\crest}{\textsc{CReST}\xspace}
\newcommand{\bisampling}{\textsc{Bi-Sampling}\xspace}
\newcommand{\cadr}{\textsc{CADR}\xspace}
\newcommand{\ssdkl}{\textsc{SSDKL}\xspace}
\newcommand{\lds}{\textsc{LDS}\xspace}
\newcommand{\bmse}{\textsc{BMSE}\xspace}
\newcommand{\ggnn}{\textsc{G$^2$GNN}\xspace}
\newcommand{\infograph}{\textsc{InfoGraph}\xspace}

\newcommand{\fixmatch}{\textsc{FixMatch}\xspace}
\newcommand{\mixmatch}{\textsc{MixMatch}\xspace}

\definecolor{amber}{rgb}{1.0, 0.49, 0.0}
\newcommand{\overbar}[1]{\mkern 1.5mu\overline{\mkern-1.5mu#1\mkern-1.5mu}\mkern 1.5mu}

\newcommand{\meltTemp}{{Plym-Melting}\xspace}
\newcommand{\density}{{Plym-Density}\xspace}
\newcommand{\oxygen}{{Plym-Oxygen}\xspace}

\newcommand{\esol}{{Mol-ESOL}\xspace}
\newcommand{\freesolv}{{Mol-FreeSolv}\xspace}
\newcommand{\lipo}{{Mol-Lipo}\xspace} % Lipophilicity

% superpixel
\newcommand{\age}{{Superpixel-Age}\xspace}

% evaluation
\newcommand{\manyrg}{{\textit{many-shot region}}\xspace}
\newcommand{\mediumrg}{{\textit{medium-shot region}}\xspace}
\newcommand{\fewrg}{{\textit{few-shot region}}\xspace}

% colors
\definecolor{mygreen}{RGB}{0 139 69}
\definecolor{mybox2}{RGB}{230 230 250}
\definecolor{mybox}{RGB}{255 218 185}
\definecolor{myred}{RGB}{205 38 38}
\definecolor{mycyan}{cmyk}{.3,0,0,0}

\newcommand{\cmidruleshiftedbyone}[1]{\cmidrulehelp#1\relax}
\def\cmidrulehelp#1-#2\relax{\cmidrule{\numexpr#1-1\relax-\numexpr#2-1\relax}}
\begin{abstract}
Data imbalance is easily found in annotated data when the observations of certain continuous label values are difficult to collect for regression tasks. When they come to molecule and polymer property predictions, the annotated graph datasets are often small because labeling them requires expensive equipment and effort. To address the lack of examples of rare label values in graph regression tasks, we propose a semi-supervised framework to progressively balance training data and reduce model bias via self-training. The training data balance is achieved by (1) pseudo-labeling more graphs for under-represented labels with a novel regression confidence measurement and (2) augmenting graph examples in latent space for remaining rare labels after data balancing with pseudo-labels. The former is to identify quality examples from unlabeled data whose labels are confidently predicted and sample a subset of them with a reverse distribution from the imbalanced annotated data. The latter collaborates with the former to target a perfect balance using a novel label-anchored mixup algorithm. We perform experiments in seven regression tasks on graph datasets. Results demonstrate that the proposed framework significantly reduces the error of predicted graph properties, especially in under-represented label areas.
\end{abstract}

\maketitle

\section{Introduction}\label{sec:intro}
Predicting the properties of graphs has attracted great attention from drug discovery~\cite{ramakrishnan2014quantum,wu2018moleculenet} and material design~\cite{ma2020pi1m,yuan2021imputation}, because molecules and polymers are naturally graphs. Properties such as density, melting temperature, and oxygen permeability are often in continuous value spaces~\cite{ramakrishnan2014quantum,wu2018moleculenet,yuan2021imputation}. Graph regression tasks are important and challenging. It is hard to observe label values in certain rare areas since the annotated data usually concentrate on small yet popular areas in the property spaces. Graph regression datasets are ubiquitously imbalanced. Previous attempts that address data imbalance mostly focused on categorical properties and classification tasks, however, \textit{imbalanced regression tasks on graphs are under-explored}.

Besides data imbalance, the annotated graph regression data are often small in real world. For example, measuring the property of a molecule or polymer often needs expensive experiments or simulations. It has taken nearly 70 years to collect \emph{only around 600} polymers with experimentally measured oxygen permeability in the {Polymer Gas Separation Membrane Database}~\cite{thornton2012polymer}. On the other side, we have \emph{hundreds of thousands} of unlabeled graphs.

\begin{figure*}[ht]
    \centering
    \includegraphics[width=\textwidth]{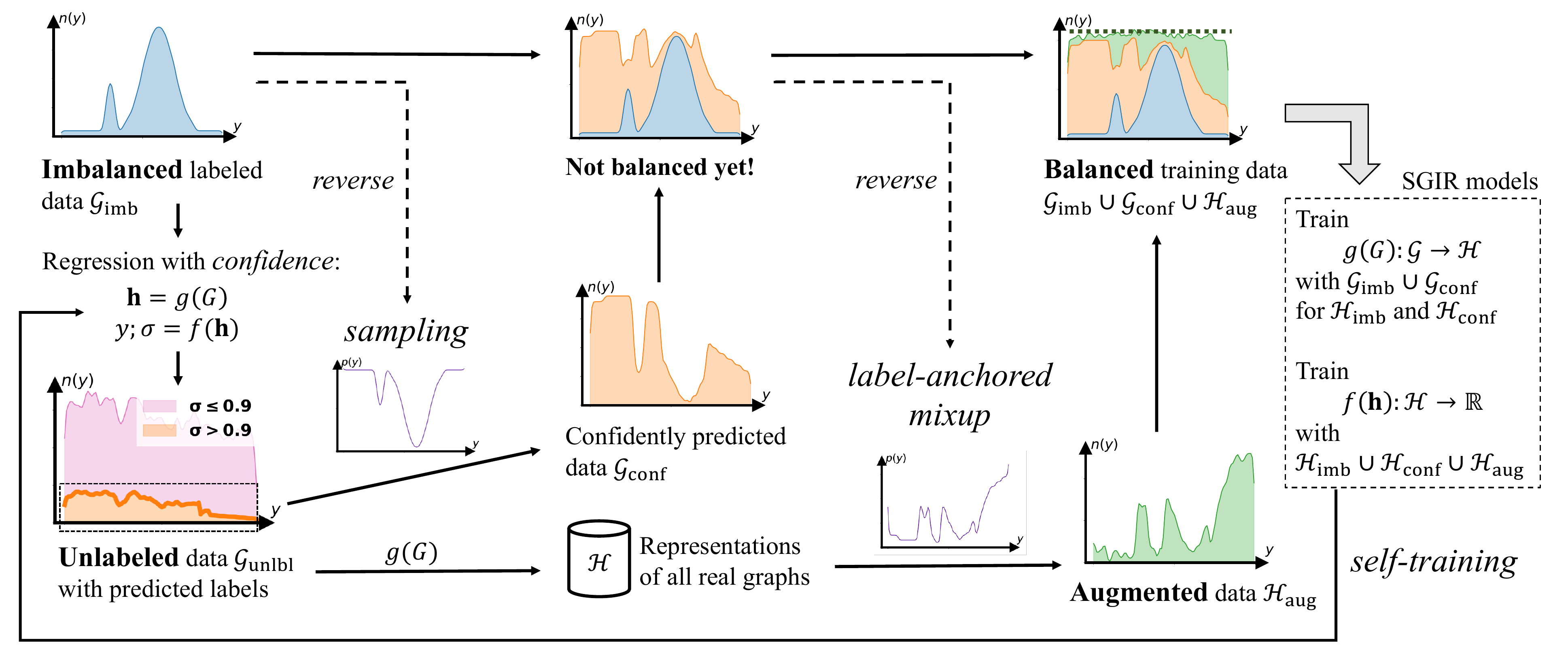}
    \vspace{-0.22in}
    \caption{An overview of our \textsc{SGIR} framework to train effective graph regression models with imbalanced labeled data. To balance the data properly, \textsc{SGIR} selects highly confident examples from predicted labels of unlabeled data and augments label areas that seriously lack data (even after added the confidently predicted data) by a novel label-anchored mixup algorithm.}
    \label{fig:sgir_framework}
    \vspace{-0.12in}
\end{figure*}

Pseudo-labeling unlabeled graphs may enrich and balance training data, however, there are two challenges.
First, if one directly trained a model on the imbalanced labeled data and used it to do pseudo-labeling, it would not be reliable to generate accurate and balanced labels.
Second, because quite a number of unlabeled graphs might not follow the distribution of labeled data, massive label noise is inevitable in pseudo-labeling and thus selection is necessary to expand the set of data examples for training. Moreover, the selected pseudo-labels without noise cannot alleviate the label imbalance problem. Because the biased model tends to generate more pseudo-labels in the label ranges where most data concentrate. In this situation, the selected pseudo-labels may aggravate the model bias and lead the model to have even worse performance on the label ranges where we lack enough data. 
Even though the pseudo-labeling had involved quality selection and the unlabeled set had been fully used to address label imbalance, the label distribution of annotated and pseudo-labeled examples might still be far from a perfect balance. This is because there might not be a sufficient number of pseudo-labeled examples to fill the gap in the under-represented label ranges.

\cref{fig:sgir_framework} illustrates our ideas to overcome the above challenges.
First, we want to progressively reduce the model bias by gradually improving training data from the labeled and unlabeled sets.
The performance of pseudo-labeling models and the quality of the expanded training data can mutually enhance each other through iterations. Second, we relate the regression confidence to the prediction variance under perturbations. Higher confidence indicates a lower prediction variance under different perturbation environments. Therefore, we define and use \emph{regression confidence} score to avoid pseudo-label noise and select quality examples in regression tasks. To fully exploit the quality pseudo-labels to compensate for the data imbalance in different label ranges, we use a reversed distribution of the imbalanced annotated data to reveal label ranges that need to be more or less selected for label balancing. 
Third, we attempt to achieve the perfect balance of training data by creating graph examples of any given label value in the remaining under-represented ranges.

In this paper, we propose \textsc{SGIR}, a novel \underline{S}emi-supervised framework for \underline{G}raph \underline{I}mbalanced \underline{R}egression. This framework has three novel designs to implement our ideas.
First, \textsc{SGIR} is a self-training framework with multiple iterations for model learning and balanced training data generation. 
Our second design is to sample more quality pseudo-labels for the less represented label ranges. We define a new measurement of regression confidence from recent studies on graph rationalization methods which provide perturbations for predictions at training and inference. After applying the confidence to filter out pseudo-label noise, we adopt \emph{reverse sampling} to find optimal sampling rates at each label value that maximize the possibility of data balance. Intuitively, if a label value is less frequent in the annotated data, the sampling rate at this value is bigger and more pseudo-labeled examples are selected for model training.
Third, we design a novel \textit{label-anchored mixup} algorithm to augment graph examples by mixing up a virtual data point and a real graph example in latent space. Each virtual point is anchored at a certain label value that is still rare in the expanded labeled data. The mixed-up graph representations continue complementing the label ranges where we seriously lack data examples.

To empirically demonstrate the advantage of \textsc{SGIR}, we conduct experiments on seven graph property regression tasks from three different domains. Results show that \textsc{SGIR} significantly reduces the prediction error on all the tasks and in both under-/well-represented label ranges. For example, on the smallest dataset \freesolv that has only 276 annotated graphs,
\textsc{SGIR} reduces the mean absolute error from 1.114 to 0.777 (relatively 30\% improvement) in the most under-represented label range and reduces the error from 0.642 to 0.563 (12\% improvement) in the entire label space compared to state-of-the-art graph regression methods. To summarize:
\begin{compactitem}
    \item We address a new problem of graph imbalance regression with a novel semi-supervised framework \method.
    \item \method is a novel self-training framework creating balanced and enriched training data from pseudo-labels and augmented examples with three collaborated components: regression confidence, reverse sampling, and label-anchored mixup.
    \item \method is theoretically motivated and empirically validated on seven graph regression tasks. It outperforms other semi-supervised learning and imbalanced regression methods in both well-represented and under-represented label ranges.
\end{compactitem}

\section{Related Work}\label{sec:related}
\subsection{Imbalanced Learning}
Data resampling is known as under-sampling majority classes or over-sampling minority classes.
\textsc{SMOTE}~\cite{chawla2002smote} created synthetic data for minority classes using linear interpolations on labeled data. Cost-sensitive techniques ~\cite{cui2019class, lin2017focal} assigned higher weights to the loss of minority classes. And posterior re-calibration~\cite{cao2019learning,tian2020posterior,menon2021longtail} encouraged larger margins for the prediction logits of minority classes. Imbalanced regression tasks have unique challenges due to continuous label values~\cite{yang2021delving}. Some of the methods from imbalanced classifications were extended to imbalanced regression tasks. 
For example, \textsc{SMOGN}~\cite{branco2017smogn} adopted the idea and method of \textsc{SMOTE} for regression; Recently, \citet{yang2021delving} used regression focal loss and cost-sensitive reweighting; and \textsc{BMSE}~\cite{ren2021bmse} used logit re-calibration to predict numerical labels. \textsc{LDS}~\cite{yang2021delving} smoothed label distribution using kernel density estimation.
\textsc{RankSim}~\cite{gong2022ranksim} regularized the latent space by approximating the distance of data points in the label space. Although these methods would improve performance on under-represented labels, they come at the expense of decreased performance on well-represented labels, particularly when annotated data is limited. \method avoids this by using unlabeled graphs to create more labels in the under-represented label ranges.

\subsection{Semi-supervised Learning}
To exploit unlabeled data, semi-supervised image classifiers such as \fixmatch~\cite{sohn2020fixmatch} and \mixmatch~\cite{berthelot2019mixmatch} used pseudo-labeling and consistency regularization. Their performance relies on weak and strong data augmentation techniques, which are under-explored for regression tasks and graph property prediction tasks. At the same time, semi-supervised learners suffer from the model bias caused by the unlabeled imbalance. Therefore, after pseudo-labeling unlabeled data, \textsc{DARP}~\cite{kim2020distribution} and \textsc{DASO}~\cite{oh2022distribution} refined the biased pseudo-labels by aligning their distribution with an approximated true class distribution of unlabeled data. \cadr~\cite{hu2022on} adjusted the threshold for pseudo-label assignments. \crest~\cite{wei2021crest} selected more pseudo-labels for minority classes in self-training. To the best of our knowledge, there was no work that leveraged unlabeled data for regression tasks on imbalanced graph data, although \textsc{SSDKL}~\cite{jean2018semi} performed semi-supervised regression for non-graph data without considering label imbalance. \method makes the first attempt to solve the imbalanced regression problem using semi-supervised learning.

\subsection{Graph Property Prediction} 
Graph neural network models (GNN)~\cite{kipf2017semi, velivckovic2018graph,hamilton2017inductive,xu2018how} have demonstrated their power for regression tasks in the fields of biology, chemistry, and material science~\cite{hu2022on,liu2022graph}. 
Data augmentation~\cite{zhao2022graph,liu2023data} is an effective way to exploit limited labeled data. The node- and link-level augmentations~\cite{rong2019dropedge,zhao2021data,zhao2022learning}
modified graph structure to improve the accuracy of node classification and link prediction. On the graph level, augmentation methods were mostly designed for classification~\cite{han2022g,wang2021mixup}. Recently, \textsc{GREA}~\cite{liu2022graph} delivered promising results for predicting polymer properties. But the model bias caused by imbalanced continuous labels was not addressed. \textsc{InfoGraph}~\cite{sun2020infograph} exploited unlabeled graphs, however, the data imbalance issue was not addressed either. Our work aims to achieve balanced training data for graph regression in real practice where we have a small set of imbalanced labeled graphs and a large set of unlabeled data.

\section{Problem Definition}\label{sec:problem}
To predict the property $y \in \mathbb{R}$ of a graph $G \in \mathcal{G}$, a graph regression model usually consists of an encoder $g: G \rightarrow \mathbf{h} \in \mathbb{R}^{d}$ and a decoder $f: \mathbf{h} \rightarrow \hat{y} \in \mathbb{R}$. The encoder $g(\cdot)$ is often a graph neural network (GNN)
that outputs the $d$-dimensional representation vector $\mathbf{h}$ of graph $G$, and the decoder $f(\cdot)$ is often a multi-layer perceptron (MLP) that makes the label prediction $\hat{y}$ given $\mathbf{h}$.

Let $\mathcal{G}_{\text{imb}} = \{(G_i, y_i) \}_{i=1}^{n_{\text{imb}}}$ denote the labeled training data for graph regression models, where $n_{\text{imb}}$ is the number of training graphs in the imbalanced labeled dataset. It often concentrates on certain areas in the continuous label space. To reveal it, we first divide the label space into $C$ intervals and use them to fully cover the range of continuous label values. These intervals are $[b_0, b_1), [b_1, b_2), \dots, [b_{C-1}, b_{C})$. Then, we assign the labeled examples into $C$ intervals and count them in each interval to construct the frequency set $\{\mu_i\}_{i=1}^{C}$. 
We could find that $\frac{ \max \{\mu_i\} }{ \min \{\mu_i\}} \gg 1 $ (\emph{i.e.}, label imbalance) often exists, instead of $\mu_1 = \mu_2 = \dots = \mu_C$ (\emph{i.e.}, label balance) that is assumed by most existing models.
The existing models may be biased to small areas in the label space that are dominated by the majority of labeled data and lack a good generalization to areas that are equally important but have much fewer examples.

Labeling continuous graph properties is difficult~\cite{yuan2021imputation}, limiting the size of labeled data. Fortunately, a large number of unlabeled graphs are often available though ignored in most existing studies. In this work, we aim to use the unlabeled examples to alleviate the label imbalance issue in graph regression tasks. That is, let $ \mathcal{G}_{\text{unlbl}} = \{G_j\}_{j=n_{\text{imb}}+1}^{n_{\text{imb}}+n_{\text{unlbl}}}$ denote the $n_{\text{unlbl}}$ available unlabeled graphs. We want to train $g(\cdot)$ and $f(\cdot)$ to deliver good performance through the whole continuous label space by utilizing both $\mathcal{G}_{\text{imb}}$ and $\mathcal{G}_{\text{unlbl}}$.

\section{Proposed Framework}\label{sec:method}
To progressively reduce label imbalance bias, we propose a novel framework named \textsc{SGIR} that iteratively creates reliable labeled examples in the areas of label space where annotations were not frequent. As presented in \cref{fig:sgir_framework}, \textsc{SGIR} uses a graph regression model to create the labels and uses the gradually balanced data to train the regression model. To let data balancing and model construction mutually enhance each other, \textsc{SGIR} is a self-training framework that trains the encoder $g(\cdot)$ and decoder $f(\cdot)$ using two strategies through multiple iterations.
The first strategy is to use pseudo-labels based on confident predictions and reverse sampling, leveraging unlabeled data (see \cref{sec:balancing_with_conf_pred}).
Because the unlabeled graph set still may not contain real examples of rare label values, the second strategy is to augment the graph representation examples for the rare areas using a novel label-anchored mixup algorithm (see \cref{sec:balancing_with_aug}).

\subsection{A Self-Training Framework for Iteratively Balancing Scalar Label Data}

A classic self-training framework is expected to be a virtuous circle exploiting the unlabeled data in label-balanced classification/regression tasks~\cite{xie2020self, mclachlan1975iterative}. It first trains a classifier/regressor that iteratively assigns pseudo-labels to the set of unlabeled training examples $\mathcal{G}_{\text{unlbl}}$ with a margin greater than a certain threshold. The pseudo-labeled examples are then used to enrich the labeled training set. And the classifier continues training with the updated training set. For a virtuous circle of model training with imbalanced labeled set  $\mathcal{G}_{\text{imb}}$, the most confident predictions on $\mathcal{G}_{\text{unlbl}}$ should be selected to compensate for the under-represented labels, as well as to enrich the dataset $\mathcal{G}_{\text{imb}}$. In each iteration, the model becomes less biased to the majority of labels. And the less biased model can make predictions of higher accuracy and confidence on the unlabeled data. Therefore, we hypothesize that model training and data balancing can mutually enhance each other.

\textsc{SGIR} is a self-training framework targeting to generalize the model performance everywhere in the continuous label space with particularly designed balanced training data from the labeled graph data $\mathcal{G}_{\text{imb}}$, confidently selected graph data $\mathcal{G}_{\text{conf}}$, and augmented representation data $\mathcal{H}_{\text{aug}}$.
For the next round of model training, the gradually balanced training data reduce the label imbalance bias carried by the graph encoder $g(\cdot)$ and decoder $f(\cdot)$.
Then the less biased graph encoder and decoder are applied to generate balanced training data of higher quality.
Through these iterations, the model bias from the imbalanced or low-quality balanced data would be progressively reduced because of the gradually enhanced quality of balanced training data.

\subsection{Balancing with Confidently Predicted Labels}\label{sec:balancing_with_conf_pred}

At each iteration, \textsc{SGIR} enriches and balances training data with pseudo-labels of good quality. The unlabeled data examples in $\mathcal{G}_{\text{unlbl}}$ are firstly exploited by reliable and confident predictions. Then the reverse sampling from the imbalanced label distribution of original training data $\mathcal{G}_{\text{imb}}$ is used to select more pseudo-labels for under-represented label ranges.

\subsubsection{Graph regression with confidence} \label{sec:graph_reg_conf}
A standard regression model outputs a scalar without a certain definition of confidence of its prediction. The confidence is often measured by how much the predicted probability is close to 1 in classifications. The lack of confidence measurements in graph regression tasks may introduce noise to the self-training framework that aims at label balancing.
It would be more severe when the domain gap exists between labeled and unlabeled data~\cite{berthelot2021adamatch}. Recent studies~\cite{liu2022graph, wu2022dir} have proposed two concepts that help us define a good measurement: rationale subgraph and environment subgraph. A rationale subgraph is supposed to best support and explain the prediction at property inference. Its counterpart environment subgraph is the complementary subgraph in the example, which perturbs the prediction from the rationale subgraph if used. Our idea is to measure the confidence of graph property prediction based on the reliability of the identified rationale subgraphs. Specifically, we use the variance of predicted label values from graphs that consist of a specific rationale subgraph and one of many possible environment subgraphs.

We use an existing {supervised} graph regression model that can identify rationale and environment subgraphs in any graph example to predict its property. We denote $G_i$ as the $i$-th graph in a batch of size $B$. The model separates $G_i$ into $G^{(r)}_i$ and $G^{(e)}_i$. For the $j$-th graph $G_j$ in the same batch, we have a combined example $G_{(i,j)} = G^{(r)}_i \cup G^{(e)}_j$ that has the rationale of $G_i$ and environment subgraph of $G_j$. So it is expected to have the same label of $G_i$. By enumerating $j \in \{1, 2, \dots, B\}$, the encoder $g(\cdot)$ and decoder $f(\cdot)$ are trained to predict the label value of any $G_{(i,j)}$.
We define the confidence of predicting the label of $G_i$ as:
\begin{equation}\label{eq:reg_conf_v1}
    \sigma_i = \frac{1}{\operatorname{Var} \left(~{\{f(g(G_{(i,j)}))\}}_{j=1,2,\dots,B}~\right)}.
\end{equation}
It is the reciprocal of prediction variance. In implementation, we choose \textsc{GREA}~\cite{liu2022graph} as the model.
% where $B$ is batch size.
Considering efficiency, \textsc{GREA} creates $G_{(i,j)}$ in the latent space without decoding its graph structure. That is, it directly gets the representation of $G_{(i,j)}$ as the sum of the representation vectors $\mathbf{h}^{(r)}_i$ of $G^{(r)}_i$ and $\mathbf{h}^{(e)}_j$ of $G^{(e)}_j$.
So we have
\begin{equation}\label{eq:reg_conf}
    \sigma_i = \frac{1}{\operatorname{Var} \left(~{\{f(\mathbf{h}^{(r)}_i+\mathbf{h}^{(e)}_j)\}}_{j=1,2,\dots,B}~\right)}.
\end{equation}
Now we have predicted labels and confidence values for graph examples in the large unlabeled dataset $\mathcal{G}_{\text{unlbl}}$.
Examples with low confidences will bring noise to the training data if we use them all. So we only consider a data example $G_i$ to be of good quality if its confidence $\sigma_i$ is not smaller than a threshold $\tau$.
We name this confidence measurement based on graph rationalization as \textsc{GRation}. \textsc{GRation} is tailored for graph regression tasks by considering the environment subgraphs as perturbations. We will compare its effect on quality graph selection against other graph-irrelevant methods such as \textsc{Dropout}~\cite{gal2016dropout}, \textsc{Certi}~\cite{tagasovska2019single}, \textsc{DER} (Deep Evidential Regression)~\cite{amini2020deep}, and \textsc{Simple} (no confidence) in experiments.

After leveraging the unlabeled data, the label distribution of quality examples may still be biased to the majority of labels.
So we further apply reverse sampling on these examples from $\mathcal{G}_{\text{unlbl}}$ to balance the distribution of training data.
 
\subsubsection{Reverse sampling}\label{sec:reverse sampling}

The reverse sampling in \textsc{SGIR} helps reduce the model bias to label imbalance. Specifically, we want to selectively add unlabeled examples predicted in the under-represented label ranges. Suppose we have the frequency set $\{\mu_i\}_{i=1}^{C}$ of $C$ intervals. We denote $p_i$ as the sampling rate at the $i$-th interval and follow~\citet{wei2021crest} to calculate it.
Basically, to perform reverse sampling, we want $p_i < p_j$ if $\mu_i > \mu_j$. We define a new frequency set $\{\mu^\prime_i\}_{i=1}^{C}$ in which $\mu^\prime_i$ equals the $i$-th smallest in $\{\mu\}$ if $\mu_i$ is the $i$-th biggest in $\{\mu\}$. Then the sampling rate is \begin{equation}\label{eq.selection_p}
    p_i = \frac{\mu^\prime_i}{\max\{\mu_1, \mu_2, \dots, \mu_C\}}.
\end{equation}
To this end, we have the confidently labeled and reversed sampled data $\mathcal{G}_{\text{conf}}$. In each self-training iteration, we combine it with the original training set $\mathcal{G}_{\text{imb}}$.

\subsection{Balancing with Augmentation via Label-Anchored Mixup}\label{sec:balancing_with_aug}
Although $\mathcal{G}_{\text{imb}}\cup\mathcal{G}_{\text{conf}}$ is more balanced than $\mathcal{G}_{\text{imb}}$, we observe that $\mathcal{G}_{\text{imb}}\cup\mathcal{G}_{\text{conf}}$ is usually far from a \emph{perfect balance}, even if $\mathcal{G}_{\text{unlbl}}$ could be hundreds of times bigger than $\mathcal{G}_{\text{imb}}$. To create graph examples targeting the remaining under-represented label ranges, we design a novel label-anchored mixup algorithm for graph imbalanced regression. Compared to existing mixup algorithms~\cite{wang2021mixup,verma2019manifold} for classifications without awareness of imbalance, our new algorithm can augment training data with additional examples for target ranges of continuous label value.

A mixup operation in the label-anchored mixup is to mix up two things in a latent space: (1) a virtual data point representing an interval of targeted label and (2) a real graph example. Specifically, we first calculate the representation of a target label interval by averaging the representation vectors of graphs in the interval from the labeled dataset $\mathcal{G}_{\text{imb}}$. Let $\mathbf{M} \in {\{0,1\}}^{C \times n_{\text{imb}}}$ be an indicator matrix, where $M_{i,j}=1$ means that the label of $G_j \in \mathcal{G}_{\text{imb}}$ belongs to the $i$-th interval. We denote $\mathbf{H} \in \mathbb{R}^{n_{\text{imb}} \times d}$ as the matrix of graph representations from the GNN encoder $g(\cdot)$ for $\mathcal{G}_{\text{imb}}$. The representation matrix $\mathbf{Z} \in \mathbb{R}^{C \times d}$ of all intervals is calculated
\begin{equation}\label{eq:interval_center_rep}
    \mathbf{Z} = \operatorname{norm}(\mathbf{M}) \cdot \mathbf{H},
\end{equation}
where $\operatorname{norm}(\cdot)$ is the row-wise normalization. Let $a_i$ denote the center label value of the $i$-th interval. 
Then we have the representation-label pairs of all the label intervals $\{(\mathbf{z}_i, a_i)\}_{i=1}^C$, where $\mathbf{z}_i$ is the $i$-th row of $\mathbf{Z}$.

Now we can use each interval center $a_i$ as a label anchor to augment graph data examples in a latent space.
We select $n_i \propto p_i$ real graphs from $\mathcal{G}_{\text{imb}}\cup\mathcal{G}_{\text{conf}}$ whose labels are closest to $a_i$, where $p_i$ is calculated by \cref{eq.selection_p}.
The more real graphs are selected, the more graph representations are augmented. $n_i$ is likely to be big when the label anchor $a_i$ remains under-represented after $\mathcal{G}_{\text{conf}}$ is added to training set.
Note that the labels were annotated if the graphs were in $\mathcal{G}_{\text{imb}}$ and predicted if they were in $\mathcal{G}_{\text{unlbl}}$.
For $j \in \{1,2,\dots,n_i\}$, we mix up the interval~($\mathbf{z}_i$, $a_i$) and a real graph ($\mathbf{h}_j$, $y_j$), where $\mathbf{h}_j$ and $y_i$ are the representation vector and the annotated or predicted label of the $j$-th graph, respectively.
Then the mixup operation is defined as
\begin{equation}\label{eq:mix_center_and_real}
    \begin{cases}
      \tilde{\mathbf{h}}_{(i,j)} &= \lambda \cdot \mathbf{z}_i + \big( 1 - \lambda \big) \cdot \mathbf{h}_j, \\
      \tilde{y}_{(i,j)} &= \lambda \cdot a_i + \big( 1 - \lambda \big) \cdot y_j, \\
    \end{cases}       
\end{equation}
where $\tilde{\mathbf{h}}_{(i,j)}$ and $\tilde{y}_{(i,j)}$ are the representation vector and label of the augmented graph, respectively. $\lambda = \max(\lambda^\prime, 1-\lambda^\prime)$, $\lambda^\prime \sim \text{Beta}(1, \beta)$, and $\beta$ is a hyperparameter. $\lambda$ is often closer to 1 because we want $\tilde{y}_{(i,j)}$ to be closer to the label anchor $a_i$.
Let $\mathcal{H}_{\text{aug}}$ denote the set of representation vectors of all the augmented graphs. Combined with $\mathcal{G}_{\text{imb}}$ and $\mathcal{G}_{\text{conf}}$, we end up with a label-balanced training set for the next round of self-training.

\subsection{Optimization}

In each iteration of self-training, we jointly optimize the parameters of graph encoder $g(\cdot)$ and label predictor $f(\cdot)$ with a gradually balanced training set $\mathcal{G}_{\text{imb}}\cup\mathcal{G}_{\text{conf}}\cup\mathcal{H}_{\text{aug}}$.

We use the mean absolute error (MAE) as the regression loss. Specifically, for each $(G,y) \in \mathcal{G}_{\text{imb}}\cup\mathcal{G}_{\text{conf}}$, the loss is $\ell_{\text{imb+conf}} = \operatorname{MAE}(f(g(G)), y)$.
Given $(\mathbf{h},y) \in \mathcal{H}_{\text{aug}}$, the loss is $\ell_{\text{aug}} = \operatorname{MAE}(f(\mathbf{h}), y)$. So the total loss for \textsc{SGIR} is
\begin{equation}
    \mathcal{L} = \sum_{(G,y) \in \mathcal{G}_{\text{imb}}\cup\mathcal{G}_{\text{conf}}} \ell_{\text{imb+conf}}(G,y) + \sum_{(\mathbf{h},y) \in \mathcal{H}_{\text{aug}}} \ell_{\text{aug}}(\mathbf{h},y). \nonumber
\end{equation}
Our framework is flexible with any graph encoder-decoder models. To be consistent and given the promising results in graph regression tasks, we use the design of graph encoder and decoder in \textsc{GREA}~\cite{liu2022graph} which is also used for measuring prediction confidence in~\cref{eq:reg_conf}.

\begin{figure}[t]
    \centering
    {\includegraphics[width=0.99\linewidth]{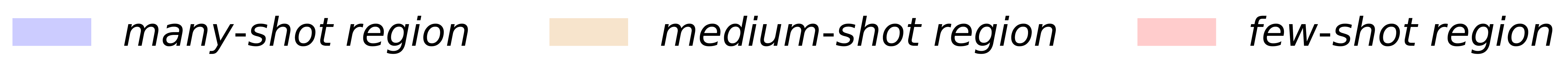}
    \label{fig:region_legend_for_train}}
    {\includegraphics[width=0.98\linewidth]{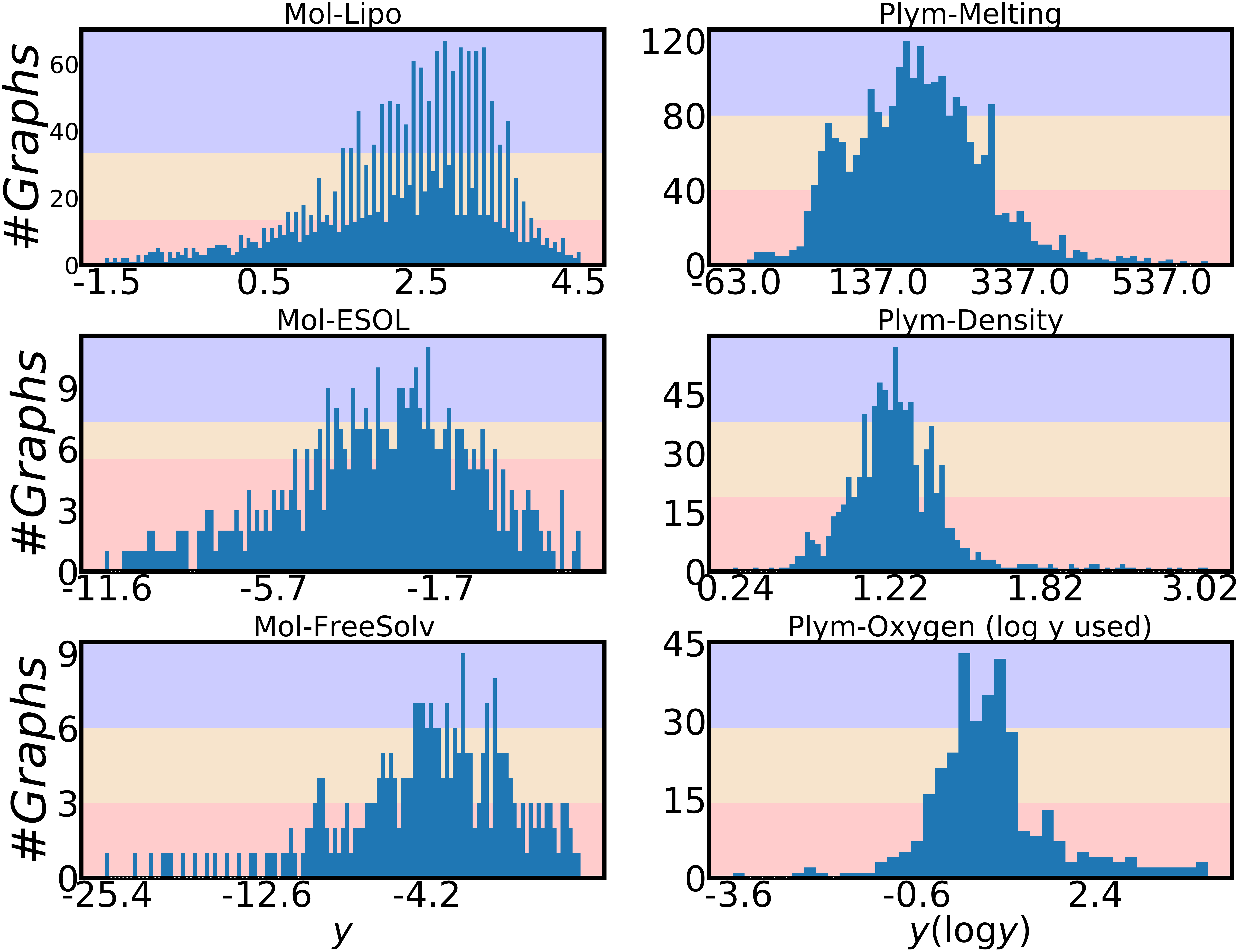}
    \label{fig:six_training_dist}}
    \vspace{-0.1in}
    \caption{Imbalanced training distributions $\mathcal{G}_{\text{imb}}$ of annotated molecule and polymers.}
    \label{fig:trainset_dist}
\end{figure}

\subsection{Theoretical Motivations}
There is a lack of theoretical principle for imbalanced regression. Our theoretical motivation extends the generalization error bound from classification~\citep{cao2019learning} to regression. The original bound enforces bigger margins for minority classes, which potentially hurt the model performance for well-represented classes~\citep{tian2020posterior,zhang2023deep}. Our result provides a more safe way to reduce the error bound by utilizing unlabeled graphs with self-training in graph regression tasks.

As we divide the label distribution into $C$ intervals, every graph example can be assigned into an interval (as the ground-truth interval) according to the distance between the interval center and the ground-truth label value. Besides, we use $S_{[b_i, b_{i+1})}(G)$ to denote the reciprocal of the distance between the predicted label of the graph $G$ and the $i$-th interval $[b_i, b_{i+1})$, where $i \in \{1, 2, \dots, C\}$. In this way, we could define $f(\cdot)$ as a regression function that outputs a continuous predicted label. Then $S_{[b_i, b_{i+1})}(G)$ consists of $f(\cdot)$ and outputs the logits to classify the graph to the $i$-the interval.

We consider all training examples to follow the same distribution. We assume that conditional on label intervals, the distributions of graph sampling are the same at training and testing stages. So, the standard 0-1 test error on the balanced test distribution is
\begin{equation}
   \mathcal{E}_\textup{bal} \left[f \right] =  \underset{\left( G, [b_i, b_{i+1}) \right) \sim \mathcal{P}_\textup{bal}}{\operatorname{Pr}} \left[ S_{[b_i, b_{i+1})}(G) < \max_{j \neq i}  S_{[b_j, b_{j+1})}(G) \right],
\end{equation}
where $\mathcal{P}_\textup{bal}$ denotes the balanced test distribution. It first samples a label interval uniformly and then samples graphs conditionally on the interval. The error for the $i$-th interval $[b_i, b_{i+1})$  is defined as 
\begin{equation}\label{eq: margin of an example}
\mathcal{E}_{[b_i, b_{i+1})} \left[f \right] = \underset{G \sim \mathcal{P}_{[b_i, b_{i+1})}}{\operatorname{Pr}} \left[S_{[b_i, b_{i+1})}(G) < \max_{j \neq i}  S_{[b_j, b_{j+1})}(G) \right],
\end{equation}
where $\mathcal{P}_{[b_i, b_{i+1})}$ denotes the distribution for the interval ${[b_i, b_{i+1})}$. We define $\gamma \left(G, [b_i, b_{i+1})\right) = S_{[b_i, b_{i+1})}(G) -  \max_{j \neq i}  S_{[b_j, b_{j+1})}(G) $ as the margin of an example $G$ assigned to the interval $[b_i, b_{i+1})$. To define the training margin $\gamma_{_{[b_i, b_{i+1})}}$ for the interval $[b_i, b_{i+1})$, we calculate the minimal margin across all examples assigned to that interval:
\begin{equation}\label{eq: margin of an interval}
    \gamma_{_{[b_i, b_{i+1})}} = \min_{G_j \in [b_i, b_{i+1})} \gamma \left( G_j,[b_i, b_{i+1}) \right).
\end{equation}

We assume that the $\operatorname{MAE}$ regression loss is small enough to correctly assign all training examples to the corresponding intervals. Given the hypothesis class $\mathcal{F}$, $\textup{C}(\mathcal{F})$ is assumed to be a proper complexity measure of $\mathcal{F}$. We assume there are $n_{[b_i, b_{i+1})}$ examples i.i.d sampled from the conditional distribution $\mathcal{P}_{[b_i, b_{i+1})}$ for the interval $[b_i, b_{i+1})$. So, we apply the standard margin-based generalization bound to obtain the following theorem~\cite{kakade2008complexity,cao2019learning,zhao2021synergistic}:
\begin{theorem}\label{theo:motivation} With probability ($1-\delta$) over the randomness of the training data, the error $\mathcal{E}_{[b_i, b_{i+1})}$ for interval $[b_i, b_{i+1})$ is bounded by 
    \begin{equation}\label{eq:reg imbalance error bound interval}
    \begin{split}
        \mathcal{E}_{[b_i, b_{i+1})}[f] & \lessapprox
        \frac{1}{\gamma_{_{[b_i, b_{i+1})}}}  \sqrt{\frac{\textup{C}(\mathcal{F}) }{n_{[b_i, b_{i+1})}}} \\
        & + \sqrt{\frac{\log\log_2(1/\gamma_{_{[b_i, b_{i+1})}}) + \log (1/\delta)}{n_{[b_i, b_{i+1})}}},
    \end{split}
    \end{equation}
    where $\lessapprox$ hides constant terms. Taking union bound over all intervals, we have $\mathcal{E}_{\textup{bal}}[f] \lessapprox \frac{1}{C}\sum_{i=1}^C \mathcal{E}_{[b_i, b_{i+1})}[f] $.
\end{theorem}
Proofs are in~\cref{add:sec:proof}. The bound decreases as the increase of the examples in corresponding label ranges. The \method is motivated to reduce and balance the bound for different intervals by manipulating $n_{[b_i, b_{i+1})}$ with pseudo-labels and augmented examples. Particularly, we discuss in~\cref{add:sec:theorem_discussion} that the augmented examples do not break our assumption for the theorem and future directions of imbalanced regression theories without intervals.

\section{Experiments}\label{sec:exp}
\begin{table}[t]
    \caption{Statistics of six tasks for graph property regression.}
    \label{tab:dataset_stat}
    \vspace{-0.1in}
    \centering
    \renewcommand{\arraystretch}{1.15}
    \renewcommand{\tabcolsep}{0.5mm}
\resizebox{0.98\linewidth}{!}{
\begin{tabular}{lllll}
\toprule
Dataset & \# Graphs \scriptsize{(Train/Valid/Test)} & \# Nodes \scriptsize{(Avg./Max)} & \# Edges \scriptsize{(Avg./Max)} \\
\midrule
\lipo & 2,048 / 1,076 / 1,076 & 27.0 / 115 & 59.0 / 236 \\
\esol & 446 / 341 / 341 & 13.3 / 55 & 27.4 / 125 \\
\freesolv & 276 / 183 / 183 & 8.7 / 24 &   16.8 / 50 \\
\midrule
\meltTemp & 2,419 / 616 / 616 & 26.9 / 102 & 55.4 / 212 \\
\density & 844 / 425 / 425 & 27.3 / 93 & 57.6 / 210 \\
\oxygen & 339 / 128 / 128 & 37.3 / 103 & 82.1 / 234 \\
\midrule
\age & 3619 / 628 / 628 & 67.9 / 75.0 & 265.6 / 300 \\
\bottomrule
\end{tabular}
}
\end{table}

\begin{table*}[ht!]
\centering
\caption{Results of \textsc{Mean\scriptsize(Std)} on six molecule/polymer datasets. The best mean is \textbf{bolded}. The best baseline is \underline{underlined}.}
\vspace{-0.1in}
\label{table:six_dataset_results}
\setlength{\tabcolsep}{4pt}
\resizebox{0.95\linewidth}{!}{\begin{tabular}{@{}llccccccccc@{}}\toprule[1.2pt]
\multicolumn{2}{c}{} & \multicolumn{4}{c}{{MAE $\downarrow$}} & \phantom{abc}& \multicolumn{4}{c}{{GM $\downarrow$}} \\
\cmidrule{3-6} \cmidrule{8-11} 
\multicolumn{2}{c}{} & All & Many-shot & Med.-shot & Few-shot &&  All & Many-shot & Med.-shot & Few-shot \\
\midrule[1.2pt]
\multirow{7}{*}{\lipo}
& \textsc{GNN} 
& 0.485{\scriptsize(0.010)} & 0.421{\scriptsize(0.030)} & 0.462{\scriptsize(0.013)} & 0.566{\scriptsize(0.032)} && 0.297{\scriptsize(0.012)} & 0.252{\scriptsize(0.022)} & 0.294{\scriptsize(0.016)} & \underline{0.348}{\scriptsize(0.030)} \\
& \textsc{RankSim}
& 0.475{\scriptsize(0.018)} & \underline{0.388}{\scriptsize(0.017)} & 0.438{\scriptsize(0.007)} & 0.587{\scriptsize(0.043)} && 0.297{\scriptsize(0.015)} & \underline{0.249}{\scriptsize(0.017)} & \underline{0.274}{\scriptsize(0.006)} & 0.380{\scriptsize(0.044)} \\
& \textsc{BMSE} 
& 0.494{\scriptsize(0.007)} & 0.409{\scriptsize(0.019)} & 0.450{\scriptsize(0.007)} & 0.614{\scriptsize(0.033)} && 0.304{\scriptsize(0.008)} & 0.260{\scriptsize(0.014)} & 0.279{\scriptsize(0.015)} & 0.382{\scriptsize(0.038)} \\
& \textsc{LDS}
& \underline{0.468}{\scriptsize(0.009)} & 0.394{\scriptsize(0.012)} & 0.449{\scriptsize(0.012)} & \underline{0.551}{\scriptsize(0.026)} && 0.294{\scriptsize(0.010)} & 0.251{\scriptsize(0.009)} & 0.281{\scriptsize(0.010)} & 0.356{\scriptsize(0.033)} \\
& \textsc{InfoGraph}
& 0.499{\scriptsize(0.008)} & 0.421{\scriptsize(0.024)} & 0.471{\scriptsize(0.013)} & 0.596{\scriptsize(0.026)} && 0.314{\scriptsize(0.011)} & 0.269{\scriptsize(0.018)} & 0.300{\scriptsize(0.006)} & 0.376{\scriptsize(0.029)} \\
& \textsc{GREA}     
& 0.487{\scriptsize(0.002)} & 0.391{\scriptsize(0.015)} & \underline{0.434}{\scriptsize(0.008)} & 0.626{\scriptsize(0.018)} && \underline{0.294}{\scriptsize(0.010)} & 0.251{\scriptsize(0.009)} & 0.281{\scriptsize(0.010)} & 0.356{\scriptsize(0.033)} \\
& \textsc{SGIR}    
& \textbf{0.432}{\scriptsize(0.012)} & \textbf{0.357}{\scriptsize(0.019)} & \textbf{0.413}{\scriptsize(0.017)} & \textbf{0.515}{\scriptsize(0.020)} && \textbf{0.264}{\scriptsize(0.013)} & \textbf{0.224{}\scriptsize(0.016)} & \textbf{0.256}{\scriptsize(0.017)} & \textbf{0.314}{\scriptsize(0.015)} \\
\midrule

\multirow{7}{*}{\esol}
& \textsc{GNN}
& 0.508{\scriptsize(0.015)} & 0.398{\scriptsize(0.018)} & 0.448{\scriptsize(0.012)} & 0.696{\scriptsize(0.025)} && 
0.299{\scriptsize(0.017)} & 0.231{\scriptsize(0.017)} & 
0.279{\scriptsize(0.014)} & 0.425{\scriptsize(0.035)} \\
& \textsc{RankSim}
& 0.501{\scriptsize(0.014)} & \underline{0.389}{\scriptsize(0.021)} & \underline{0.443}{\scriptsize(0.019)} & 0.689{\scriptsize(0.025)} && 0.293{\scriptsize(0.021)} & 0.227{\scriptsize(0.028)} & \underline{0.258}{\scriptsize(0.020)} & 0.449{\scriptsize(0.030)} \\
& \textsc{BMSE} 
& 0.533{\scriptsize(0.023)} & 0.400{\scriptsize(0.027)} & 0.449{\scriptsize(0.015)} & 0.777{\scriptsize(0.069)} && 0.308{\scriptsize(0.018)} & 0.245{\scriptsize(0.036)} & 0.266{\scriptsize(0.009)} & 0.473{\scriptsize(0.035)} \\
& \textsc{LDS}
& 0.517{\scriptsize(0.016)} & 0.423{\scriptsize(0.012)} & 0.474{\scriptsize(0.029)} & 0.668{\scriptsize(0.010)} &&
0.304{\scriptsize(0.010)} & 0.261{\scriptsize(0.007)} & 0.283{\scriptsize(0.025)} & \underline{0.393}{\scriptsize(0.009)} \\
& \textsc{InfoGraph}
& 0.561{\scriptsize(0.025)} & 0.475{\scriptsize(0.034)} & 0.466{\scriptsize(0.036)} & 0.776{\scriptsize(0.036)} && 0.336{\scriptsize(0.014)} & 0.306{\scriptsize(0.022)} & 0.276{\scriptsize(0.013)} & 0.484{\scriptsize(0.029)} \\
& \textsc{GREA}     
& \underline{0.497}{\scriptsize(0.031)} & 0.396{\scriptsize(0.040)} & 0.456{\scriptsize(0.033)} & \underline{0.652}{\scriptsize(0.045)} && \underline{0.289}{\scriptsize(0.032)} & \underline{\textbf{0.226}}{\scriptsize(0.038)} & 0.270{\scriptsize(0.025)} & 0.404{\scriptsize(0.051)} \\
& \textsc{SGIR}
& \textbf{0.457}{\scriptsize(0.015)} & \textbf{0.370}{\scriptsize(0.022)} 
& \textbf{0.411}{\scriptsize(0.011)} & \textbf{0.604}{\scriptsize(0.024)}  &
& \textbf{0.263}{\scriptsize(0.016)} & \textbf{ 0.226}{\scriptsize(0.021)} 
& \textbf{0.240}{\scriptsize(0.015)} & \textbf{0.347}{\scriptsize(0.030)}  \\
\midrule

\multirow{7}{*}{\freesolv}
& \textsc{GNN}  
& 0.726{\scriptsize(0.039)} & 0.617{\scriptsize(0.061)} & 0.695{\scriptsize(0.055)} & 1.154{\scriptsize(0.082)} && 0.363{\scriptsize(0.025)} & 0.317{\scriptsize(0.027)} & 0.360{\scriptsize(0.029)} & 0.556{\scriptsize(0.073)} \\
& \textsc{RankSim}
& 0.779{\scriptsize(0.109)} & 0.764{\scriptsize(0.225)} & 0.674{\scriptsize(0.072)} & 1.220{\scriptsize(0.146)} && 0.367{\scriptsize(0.026)} & 0.396{\scriptsize(0.052)} & 0.315{\scriptsize(0.030)} & \underline{0.537}{\scriptsize(0.082)} \\
& \textsc{BMSE} 
& 0.856{\scriptsize(0.071)} & 0.809{\scriptsize(0.117)} & 0.820{\scriptsize(0.064)} & 1.122{\scriptsize(0.076)} && 0.456{\scriptsize(0.042)} & 0.426{\scriptsize(0.029)} & 0.457{\scriptsize(0.054)} & 0.552{\scriptsize(0.062)} \\
& \textsc{LDS}
& 0.809{\scriptsize(0.071)} & 0.796{\scriptsize(0.071)} & 0.737{\scriptsize(0.088)} & \underline{1.114}{\scriptsize(0.141)} && 0.443{\scriptsize(0.045)} & 0.489{\scriptsize(0.036)} & 0.387{\scriptsize(0.052)} & 0.580{\scriptsize(0.146)} \\
& \textsc{InfoGraph}
& 0.933{\scriptsize(0.042)} & 0.830{\scriptsize(0.081)} & 0.913{\scriptsize(0.030)} & 1.308{\scriptsize(0.171)} && 0.542{\scriptsize(0.048)} & 0.505{\scriptsize(0.107)} & 0.528{\scriptsize(0.038)} & 0.789{\scriptsize(0.183)} \\
& \textsc{GREA} 
& \underline{0.642}{\scriptsize(0.026)} & \underline{0.541}{\scriptsize(0.064)} & \underline{0.570}{\scriptsize(0.008)} & 1.202{\scriptsize(0.023)} && \underline{0.321}{\scriptsize(0.038)} & \underline{0.294}{\scriptsize(0.064)} & \underline{0.301}{\scriptsize(0.024)} & \underline{0.537}{\scriptsize(0.049)} \\
& \textsc{SGIR}
& \textbf{0.563}{\scriptsize(0.026)} & \textbf{0.535}{\scriptsize(0.038)}
& \textbf{0.528}{\scriptsize(0.046)} & \textbf{0.777}{\scriptsize(0.061)} &
& \textbf{0.264}{\scriptsize(0.029)} & \textbf{0.286}{\scriptsize(0.013)}
& \textbf{0.244}{\scriptsize(0.046)} & \textbf{0.304}{\scriptsize(0.078)} \\
\midrule

\multirow{7}{*}{\meltTemp}
& \textsc{GNN}  
& 41.8{\scriptsize(1.2)} & 35.5{\scriptsize(1.2)} & 33.0{\scriptsize(0.7)} & 54.7{\scriptsize(2.2)} && 23.2{\scriptsize(1.0)} & 21.3{\scriptsize(1.1)} & \underline{16.2}{\scriptsize(1.0)} & 33.4{\scriptsize(2.5)} \\
& \textsc{RankSim}
& \underline{41.1}{\scriptsize(0.9)} & 34.1{\scriptsize(0.5)} & 33.6{\scriptsize(1.1)} & 53.5{\scriptsize(1.2)} && \underline{22.6}{\scriptsize(1.1)} & 20.5{\scriptsize(0.5)} & 16.8{\scriptsize(1.0)} & \underline{31.4}{\scriptsize(2.8)} \\
& \textsc{BMSE} 
& 42.1{\scriptsize(0.7)} & 35.8{\scriptsize(1.4)} & 34.1{\scriptsize(1.3)} & 54.4{\scriptsize(1.5)} && 23.7{\scriptsize(1.2)} & 21.5{\scriptsize(1.0)} & 18.1{\scriptsize(0.5)} & 32.4{\scriptsize(3.0)} \\
& \textsc{LDS}
& 41.6{\scriptsize(0.3)} & 35.3{\scriptsize(0.9)} & 34.5{\scriptsize(1.1)} & \underline{53.2}{\scriptsize(0.8)} && 
23.2{\scriptsize(0.2)} & 20.5{\scriptsize(1.2)} & 18.3{\scriptsize(0.5)} & \underline{31.4}{\scriptsize(1.1)} \\
& \textsc{InfoGraph}
& 43.6{\scriptsize(2.8)} & 35.3{\scriptsize(2.3)} & 35.0{\scriptsize(2.3)} & 58.3{\scriptsize(4.1)} && 24.6{\scriptsize(1.9)} & 21.3{\scriptsize(1.5)} & 18.4{\scriptsize(1.5)} & 35.4{\scriptsize(4.1)} \\
& \textsc{GREA}
& 41.2{\scriptsize(0.8)} & \underline{33.3}{\scriptsize(0.5)} & \underline{32.7}{\scriptsize(0.7)} & 55.3{\scriptsize(3.0)} &&
23.4{\scriptsize(0.6)} & \underline{20.0}{\scriptsize(0.6)} & 17.3{\scriptsize(0.7)} & 34.3{\scriptsize(2.9)} \\
& \textsc{SGIR}    
& \textbf{38.9}{\scriptsize(0.7)} & \textbf{31.7}{\scriptsize(0.3)} & \textbf{31.5}{\scriptsize(1.1)} & \textbf{51.4}{\scriptsize(1.6)} && 
\textbf{21.1}{\scriptsize(1.2)} & \textbf{18.5}{\scriptsize(0.5)} & \textbf{15.9}{\scriptsize(1.4)} & \textbf{30.2}{\scriptsize(1.9)} \\
\midrule

\multirow{7}{*}{\density}
& \textsc{GNN}  
& 61.2{\scriptsize(5.4)} & 63.4{\scriptsize(18.9)} & 
46.6{\scriptsize(1.6)} & 72.0{\scriptsize(2.8)} && 
\underline{29.3}{\scriptsize(0.6)} & 29.6{\scriptsize(3.3)} & 
23.5{\scriptsize(0.9)} & 35.5{\scriptsize(2.0)} \\
& \textsc{RankSim}
& 57.5{\scriptsize(1.8)} & 55.1{\scriptsize(2.2)} & 46.3{\scriptsize(1.8)} & \underline{69.4}{\scriptsize(3.3)} && 
\underline{29.3}{\scriptsize(1.6)} & 29.9{\scriptsize(2.8)} & 23.1{\scriptsize(2.1)} & \underline{35.4}{\scriptsize(2.5)} \\
& \textsc{BMSE} 
& 61.8{\scriptsize(2.0)} & 59.1{\scriptsize(8.6)} & 
48.2{\scriptsize(2.0)} & 75.9{\scriptsize(3.5)} && 
31.9{\scriptsize(1.3)} & 31.8{\scriptsize(4.2)} & 
26.3{\scriptsize(2.2)} & 38.2{\scriptsize(3.2)} \\
\multirow{3}{*}{(scaled:$ \times 1,000$)}
& \textsc{LDS}
& 60.1{\scriptsize(2.4)} & 60.4{\scriptsize(6.2)} & 47.0{\scriptsize(1.3)} & 71.3{\scriptsize(2.5)} &&
31.5{\scriptsize(2.0)} & 33.2{\scriptsize(3.5)} & 24.4{\scriptsize(3.0)} & 38.0{\scriptsize(2.4)} \\
& \textsc{InfoGraph}
& \underline{54.9}{\scriptsize(1.7)} & \underline{46.8}{\scriptsize(1.0)} & \underline{43.0}{\scriptsize(1.9)} & 72.3{\scriptsize(3.2)} && 
\underline{29.3}{\scriptsize(1.8)} & 27.3{\scriptsize(1.4)} &
\underline{\textbf{22.6}}{\scriptsize(1.2)} & 39.2{\scriptsize(4.3)} \\
& \textsc{GREA}     
& 60.3{\scriptsize(1.9)} & 49.0{\scriptsize(4.4)} & 
48.1{\scriptsize(2.5)} & 80.7{\scriptsize(4.2)} && 
32.3{\scriptsize(1.6)} & \underline{26.7}{\scriptsize(2.7)} & 
27.2{\scriptsize(2.3)} & 44.7{\scriptsize(6.1)} \\
& \textsc{SGIR}
& \textbf{53.0}{\scriptsize(0.5)} & \textbf{45.4}{\scriptsize(1.7)} & \textbf{42.5}{\scriptsize(2.8)} & \textbf{68.6}{\scriptsize(2.6)} &&
\textbf{26.6}{\scriptsize(0.4)} & \textbf{24.0}{\scriptsize(2.2)} & 23.0{\scriptsize(1.3)} & \textbf{33.4}{\scriptsize(3.0)} \\
\midrule

\multirow{7}{*}{\oxygen}
& \textsc{GNN}  
& 183.5{\scriptsize(33.4)} & 6.3{\scriptsize(3.2)} & 14.6{\scriptsize(6.6)} & 464.0{\scriptsize(85.3)} && 7.0{\scriptsize(1.8)} & 2.4{\scriptsize(0.7)} & 3.9{\scriptsize(1.1)} & 29.9{\scriptsize(7.2)} \\
& \textsc{RankSim}
& \underline{165.7}{\scriptsize(27.4)} & \underline{3.9}{\scriptsize(1.4)} & 13.0{\scriptsize(2.0)} & \underline{420.7}{\scriptsize(69.7)} && \underline{5.9}{\scriptsize(1.4)} & \underline{\textbf{1.8}}{\scriptsize(0.3)} & \underline{3.6}{\scriptsize(1.7)} & \underline{26.6}{\scriptsize(6.7)} \\
& \textsc{BMSE} 
& 190.4{\scriptsize(33.4)} & 26.4{\scriptsize(21.6)} & 27.0{\scriptsize(16.4)} & 454.3{\scriptsize(88.9)} && 25.7{\scriptsize(14.8)} & 14.9{\scriptsize(11.7)} & 15.9{\scriptsize(9.6)} & 63.2{\scriptsize(23.5)} \\
& \textsc{LDS}
& 180.0{\scriptsize(23.0)} & 6.6{\scriptsize(4.0)} & \underline{\textbf{11.8}}{\scriptsize(2.0)} & 456.3{\scriptsize(60.2)} && 7.6{\scriptsize(1.6)} & 2.4{\scriptsize(0.6)} & 4.7{\scriptsize(1.4)} & 33.6{\scriptsize(9.2)} \\
& \textsc{InfoGraph}
& 199.5{\scriptsize(31.5)} & 7.5{\scriptsize(7.2)} & 13.0{\scriptsize(1.8)} & 505.5{\scriptsize(78.2)} && 7.8{\scriptsize(1.9)} & 2.3{\scriptsize(0.5)} & 5.1{\scriptsize(2.2)} & 34.8{\scriptsize(8.5)} \\
& \textsc{GREA}     
& 182.5{\scriptsize(30.0)} & 9.0{\scriptsize(8.6)} & 14.4{\scriptsize(4.9)} & 458.8{\scriptsize(79.2)} && 7.1{\scriptsize(1.3)} & 2.1{\scriptsize(0.5)} & 4.4{\scriptsize(1.3)} & 31.7{\scriptsize(5.0)} \\
& \textsc{SGIR}    
& \textbf{150.9}{\scriptsize(17.8)} & \textbf{3.8}{\scriptsize(1.1)} & 
12.2{\scriptsize(0.6)} & 
\textbf{382.8}{\scriptsize(46.9)} && 
\textbf{5.8}{\scriptsize(0.4)} & 
2.1{\scriptsize(0.7)} & 
\textbf{3.3}{\scriptsize(0.8)} & \textbf{24.4}{\scriptsize(6.8)} \\
\bottomrule[1.5pt]
\end{tabular}}
\vspace{-0.15in}
\end{table*}

We conduct experiments to demonstrate the effectiveness of \textsc{SGIR} and answer the research question: how it performs on graph regression tasks and at different label ranges (RQ1). We also make a few ablation studies to investigate the effect of model design: where the effectiveness comes from (RQ2).

\subsection{Experimental Settings}
\subsubsection{Datasets}

\begin{figure*}[t!]

\begin{minipage}{0.2\linewidth}
    \centering
    \includegraphics[width=0.99\linewidth]{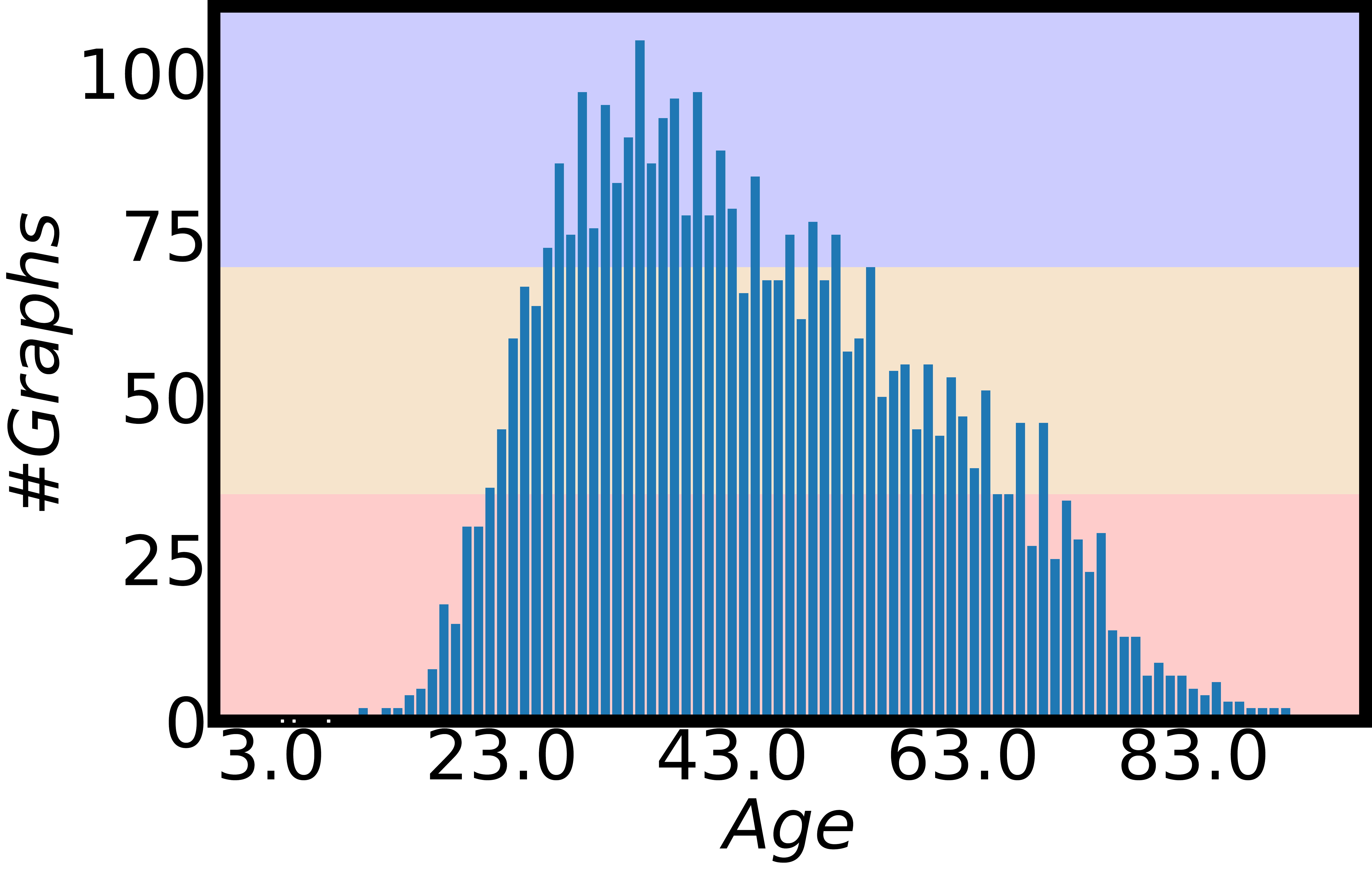}
    \vspace{-0.3in}
    \caption{\small Imbalanced training distributions $\mathcal{G}_{\text{imb}}$ in the \age dataset.}
    \label{fig:age_distribution}
\end{minipage}
\hfill
\begin{minipage}{0.79\linewidth}
\renewcommand{\arraystretch}{1.15}
\renewcommand{\tabcolsep}{0.5mm}
\vspace{-0.1in}
\resizebox{0.95\linewidth}{!}{
\begin{tabular}{@{}llccccccccc@{}}\toprule[1.2pt]
    \multicolumn{2}{c}{} & \multicolumn{4}{c}{{MAE $\downarrow$}} & \phantom{abc}& \multicolumn{4}{c}{{GM $\downarrow$}} \\
    \cmidrule{3-6} \cmidrule{8-11} 
    \multicolumn{2}{c}{} & All & Many-shot & Med.-shot & Few-shot &&  All & Many-shot & Med.-shot & Few-shot \\
    \midrule[1.2pt]
    \multirow{7}{*}{}
    & \textsc{GNN} 
    & 14.583{\scriptsize(0.413)} & 10.524{\scriptsize(0.994)} & 11.698{\scriptsize(0.404)} & 22.127{\scriptsize(0.780)} && 9.996{\scriptsize(0.386)} & 7.265{\scriptsize(0.858)} & 7.910{\scriptsize(0.492)} & 18.404{\scriptsize(0.673)} \\
    & \textsc{RankSim}
    & \underline{14.464}{\scriptsize(0.401)} & 10.468{\scriptsize(0.759)} & 11.610{\scriptsize(0.774)} & 21.910{\scriptsize(0.700)}  && \underline{9.606}{\scriptsize(0.303)} & \underline{6.936}{\scriptsize(0.598)} & 7.721{\scriptsize(0.660)} & 17.534{\scriptsize(1.768)}  \\
    & \textsc{BMSE} 
    & 15.179{\scriptsize(0.594)} & 10.639{\scriptsize(2.303)} & 12.201{\scriptsize(0.900)} & 23.321{\scriptsize(2.525)} && 10.419{\scriptsize(0.393)} & 7.249{\scriptsize(1.526)} & 8.659{\scriptsize(0.827)} & 19.719{\scriptsize(4.318)} \\
    & \textsc{LDS}
    & 14.674{\scriptsize(0.191)} & 10.972{\scriptsize(0.495)} & 11.985{\scriptsize(0.627)} & \underline{21.623}{\scriptsize(0.926)} && 9.867{\scriptsize(0.291)} & 7.317{\scriptsize(0.672)} & 7.997{\scriptsize(0.633)} & \underline{17.298}{\scriptsize(0.957)} \\
    & \textsc{InfoGraph}
    & 14.515{\scriptsize(0.605)} & 10.610{\scriptsize(1.063)} & \underline{11.150}{\scriptsize(0.158)} & 22.476{\scriptsize(1.147)} && 9.879{\scriptsize(0.524)} & 7.391{\scriptsize(0.995)} & \underline{7.377}{\scriptsize(0.333)} & 18.969{\scriptsize(1.873)} \\
    & \textsc{GREA}     
    & 14.682{\scriptsize(0.300)} & \underline{10.283}{\scriptsize(0.503)} & 11.999{\scriptsize(0.585)} & 22.329{\scriptsize(0.570)}  && 10.037{\scriptsize(0.438)} & 7.051{\scriptsize(0.455)} & 8.273{\scriptsize(0.565)} & 18.142{\scriptsize(1.276)} \\
    & \textsc{SGIR}
    & \textbf{13.787}{\scriptsize(0.123)} & \textbf{10.171}{\scriptsize(0.4156)} & \textbf{11.066}{\scriptsize(0.389)} & \textbf{20.687}{\scriptsize(0.839)} && \textbf{9.261}{\scriptsize(0.221)} & \textbf{6.928}{\scriptsize(0.355)} & \textbf{7.247}{\scriptsize(0.593)} & \textbf{16.769}{\scriptsize(1.418)} \\
    \bottomrule[1.5pt]
\end{tabular}
}
\captionof{table}{\small Results of \textsc{ Mean\scriptsize(Std)} on the age prediction using graphs from image superpixels. \\ The best mean is \textbf{bold}. The best baseline is \underline{underlined}.}
\label{table:agedb_results}
\vspace{-0.15in}
\end{minipage}
\end{figure*}

\textbf{Molecule and polymer datasets} in~\cref{tab:dataset_stat} (the datasets with a prefix Mol- or Plym-) and \cref{fig:trainset_dist} present detailed data statistics for six graph regression tasks from chemistry and materials science. Three molecule datasets are from~\cite{wu2018moleculenet} and three polymer datasets are from~\cite{liu2022graph}. Besides labeled graphs, we combine a database of 133,015 molecules in QM9~\cite{ramakrishnan2014quantum} and an integration of four sets of 13,114 polymers in total~\cite{liu2022graph} to create a set of \textbf{146,129} unlabeled graphs to set up semi-supervised graph regression.
We note that the unlabeled graphs may be slightly less than 146,129 for a polymer task on \meltTemp, \density or \oxygen. Because we remove the overlapping graph examples for the current polymer task with the polymer unlabeled data. We follow \cite{yang2021delving} to split the datasets to characterize imbalanced training distributions and balanced test distributions. The details of the \textbf{age regression dataset} are presented in~\cref{tab:dataset_stat} (\age) and~\cref{fig:age_distribution}. The graph dataset \age is constructed from image superpixels using the algorithms from~\cite{knyazev2019understanding} on the image dataset \emph{AgeDB-DIR} from~\cite{moschoglou2017agedb, yang2021delving}. Each face image in \emph{AgeDB-DIR} has an age label from 0 to 101. We fisrt compute the SLIC superpixels for each image without losing the label-specific information~\cite{achanta2012slic,knyazev2019understanding}. Then we use the superpixels as nodes and calculate the spatial distance between superpixels to build edges for each image~\cite{knyazev2019understanding}. Binary edges are constructed between superpixel nodes by applying a threshold on the top-5\% of the smallest spatial distances. After building a graph for each image, the graph dataset \age consists of 3,619 graphs for training, 628 graphs for validation, 628 graphs for testing, and 11,613 unlabeled graphs for semi-supervised learning.

\subsubsection{Evaluation metrics} We report model performance on three different sub-ranges following the work in~\cite{yang2021delving,ren2021bmse,gong2022ranksim}, besides the \emph{entire range} of label space. The three sub-ranges are the \manyrg, \mediumrg, and \fewrg.
The sub-ranges are defined by the number of training graphs in each label value interval. Details for each dataset are presented in~\cref{fig:trainset_dist} and~\cref{fig:age_distribution}. To evaluate the regression performance, we use mean absolute error (MAE) and geometric mean (GM)~\cite{yang2021delving}. Lower values ($\downarrow$) of MAE or GM indicate better performance. 
\begin{table*}
  \mbox{}\hfill
\renewcommand{\arraystretch}{0.99}
\renewcommand{\tabcolsep}{0.5mm}
\begin{minipage}[t]{.48\linewidth}
\centering
\caption{\small  A comprehensive ablation study on molecule regression datasets with the metric {MAE ($\downarrow$)}. $\sigma$ is the confidence score in~\cref{sec:graph_reg_conf}. $p$ is the reverse sampling in~\cref{sec:reverse sampling}. ($\Tilde{\mathbf{h}}$,~$\Tilde{y}$) is the label-anchored mixup in~\cref{sec:balancing_with_aug}.}
\vspace{-0.1in}
\label{table:ablation}
\resizebox{0.9\linewidth}{!}{\begin{tabular}
{@{}c|ccc|cccc@{}}\toprule[1.2pt]
 \phantom{aa} &  \phantom{aa}$ \sigma$\phantom{a}  & \phantom{a}$p$\phantom{a} & ($\Tilde{\mathbf{h}}$, $\Tilde{y}$) & All & Many-shot & Med.-shot & Few-shot  \\
\midrule[1.2pt]
\parbox[t]{2mm}{\multirow{5}{*}{\rotatebox[origin=c]{90}{\lipo}}} 
& \multicolumn{3}{c|}{w/o $\mathcal{G}_\text{unlbl}$}
& 0.477{\scriptsize(0.014)} & 0.378{\scriptsize(0.030)} 
& 0.440{\scriptsize(0.011)} & 0.600{\scriptsize(0.006)} \\
\cmidrule{2-8} 
& \cmark & \xmark & \xmark
& 0.448{\scriptsize(0.006)} & 0.371{\scriptsize(0.004)} 
& 0.421{\scriptsize(0.012)} & 0.543{\scriptsize(0.016)} \\
& \xmark & \cmark & \xmark
& 0.446{\scriptsize(0.008)} & \textbf{0.356}{\scriptsize(0.003)} 
& \textbf{0.407}{\scriptsize(0.011)} & 0.564{\scriptsize(0.016)} \\
& \cmark & \cmark & \xmark
& 0.442{\scriptsize(0.012)} & 0.372{\scriptsize(0.007)} 
& 0.415{\scriptsize(0.004)} & 0.533{\scriptsize(0.026)} \\
& \xmark & \xmark & \cmark
& 0.456{\scriptsize(0.007)} & 0.372{\scriptsize(0.014)} 
& 0.436{\scriptsize(0.010)} & 0.549{\scriptsize(0.005)} \\
& \cmark & \cmark & \cmark
& \textbf{0.432}{\scriptsize(0.012)} & 0.357{\scriptsize(0.019)} 
& 0.413{\scriptsize(0.017)} & \textbf{0.515}{\scriptsize(0.020)} \\

\midrule
\parbox[t]{2mm}{\multirow{5}{*}{\rotatebox[origin=c]{90}{\esol}}} 
& \multicolumn{3}{c|}{w/o $\mathcal{G}_\text{unlbl}$}
& 0.477{\scriptsize(0.027)} & 0.375{\scriptsize(0.014)} 
& 0.432{\scriptsize(0.042)} & 0.637{\scriptsize(0.042)} \\
\cmidrule{2-8} 
& \cmark & \xmark & \xmark
& 0.475{\scriptsize(0.014)} & 0.369{\scriptsize(0.014)} 
& 0.446{\scriptsize(0.017)} & 0.618{\scriptsize(0.039)} \\
& \xmark & \cmark & \xmark
& 0.480{\scriptsize(0.017)} & 0.380{\scriptsize(0.035)} 
& 0.440{\scriptsize(0.017)} & 0.630{\scriptsize(0.020)} \\
& \cmark & \cmark & \xmark
& 0.468{\scriptsize(0.007)} & 0.379{\scriptsize(0.012)} 
& 0.425{\scriptsize(0.013)} & 0.612{\scriptsize(0.028)} \\
& \xmark & \xmark & \cmark
& 0.474{\scriptsize(0.010)} & \textbf{0.353}{\scriptsize(0.018)} 
& 0.450{\scriptsize(0.009)} & 0.623{\scriptsize(0.027)} \\
& \cmark & \cmark & \cmark
& \textbf{0.457}{\scriptsize(0.015)} & 0.370{\scriptsize(0.022)} 
& \textbf{0.411}{\scriptsize(0.011)} & \textbf{0.604}{\scriptsize(0.024)} \\
\midrule
\parbox[t]{2mm}{\multirow{5}{*}{\rotatebox[origin=c]{90}{\freesolv}}} 
& \multicolumn{3}{c|}{w/o $\mathcal{G}_\text{unlbl}$}
& 0.619{\scriptsize(0.019)} & \textbf{0.525}{\scriptsize(0.022)} 
& 0.590{\scriptsize(0.035)} & 1.000{\scriptsize(0.072)} \\
\cmidrule{2-8} 
& \cmark & \xmark & \xmark
& 0.604{\scriptsize(0.020)} & 0.557{\scriptsize(0.037)} 
& 0.560{\scriptsize(0.029)} & 0.903{\scriptsize(0.055)} \\
& \xmark & \cmark & \xmark
& 0.660{\scriptsize(0.028)} & 0.574{\scriptsize(0.015)} 
& 0.650{\scriptsize(0.036)} & 0.941{\scriptsize(0.066)} \\
& \cmark & \cmark & \xmark
& 0.568{\scriptsize(0.029)} & 0.538{\scriptsize(0.020)} 
& \textbf{0.520}{\scriptsize(0.045)} & 0.831{\scriptsize(0.132)} \\
& \xmark & \xmark & \cmark
& 0.593{\scriptsize(0.045)} & 0.536{\scriptsize(0.033)} 
& 0.542{\scriptsize(0.067)} & 0.947{\scriptsize(0.062)} \\
& \cmark & \cmark & \cmark
& \textbf{0.563}{\scriptsize(0.026)} & 0.535{\scriptsize(0.038)}
& 0.528{\scriptsize(0.046)} & \textbf{0.777}{\scriptsize(0.061)} \\

\bottomrule[1.5pt]
\end{tabular}}
\end{minipage}\hfill
\begin{minipage}[t]{0.48\linewidth}
\renewcommand{\arraystretch}{1.18}
\centering
\caption{\small Investigation on choices of regression confidence with the metric {MAE ($\downarrow$)}. We disable all other \method components except the regression confidence score. Our confidence score (\textbf{GRation}) in~\cref{eq:reg_conf_v1} removes noise more effectively than others in graph regression tasks. }
\vspace{-0.1in}
\label{table:uncertainty}
\resizebox{0.9\linewidth}{!}{\begin{tabular}{@{}l|l|clll@{}}\toprule[1.2pt]
\phantom{aa}& Choice of $\sigma$ & All & Many-shot & Med.-shot & Few-shot \\
\midrule[1.2pt]
\parbox[t]{2mm}{\multirow{5}{*}{\rotatebox[origin=c]{90}{\lipo}}} 
& \textsc{Simple}  & 
0.481{\scriptsize(0.010)} & 0.389{\scriptsize(0.007)} & 0.440{\scriptsize(0.013)} & 0.603{\scriptsize(0.023)} \\
& \textsc{Dropout} & 0.450{\scriptsize(0.026)} & \textbf{0.365}{\scriptsize(0.031)} & \textbf{0.420}{\scriptsize(0.022)} & 0.555{\scriptsize(0.037)} \\
& \textsc{Certi} & 
0.452{\scriptsize(0.011)} & 0.384{\scriptsize(0.018)} & 0.433{\scriptsize(0.013)} & \textbf{0.532}{\scriptsize(0.010)} \\
& \textsc{DER} & 
1.026{\scriptsize(0.033)} & 0.604{\scriptsize(0.035)} & 0.760{\scriptsize(0.016)} & 1.672{\scriptsize(0.111)} \\
& \textsc{GRation} & 
\textbf{0.448}{\scriptsize(0.006)} & 0.371{\scriptsize(0.004)} & 0.421{\scriptsize(0.012)} & 0.543{\scriptsize(0.016)} \\
\midrule
\parbox[t]{2mm}{\multirow{5}{*}{\rotatebox[origin=c]{90}{\esol}}} 
& \textsc{Simple}  &
0.499{\scriptsize(0.016)} & 0.397{\scriptsize(0.023)} & 0.457{\scriptsize(0.018)} & 0.656{\scriptsize(0.033)}  \\
& \textsc{Dropout} & 
0.483{\scriptsize(0.011)} & 0.381{\scriptsize(0.027)} & \textbf{0.443}{\scriptsize(0.018)} & 0.636{\scriptsize(0.027)} \\
& \textsc{Certi} & 
0.487{\scriptsize(0.030)} & 0.389{\scriptsize(0.039)} & 0.439{\scriptsize(0.024)} & 0.647{\scriptsize(0.043)} \\
& \textsc{DER} & 
0.918{\scriptsize(0.135)} & 0.776{\scriptsize(0.086)} & 0.826{\scriptsize(0.098)} & 1.182{\scriptsize(0.245)} \\
& \textsc{GRation} & 
\textbf{0.475}{\scriptsize(0.014)} & \textbf{0.369}{\scriptsize(0.014)} & 0.446{\scriptsize(0.017)} & \textbf{0.618}{\scriptsize(0.039)} \\
\midrule
\parbox[t]{2mm}{\multirow{5}{*}{\rotatebox[origin=c]{90}{\freesolv}}} 
& \textsc{Simple} &
0.697{\scriptsize(0.056)} & 0.616{\scriptsize(0.025)} & 0.663{\scriptsize(0.033)} & 1.054{\scriptsize(0.260)} \\
& \textsc{Dropout} & 
0.639{\scriptsize(0.013)} & 0.578{\scriptsize(0.060)} & 0.589{\scriptsize(0.017)} & 1.005{\scriptsize(0.140)} \\
& \textsc{Certi} & 
0.654{\scriptsize(0.049)} & 0.589{\scriptsize(0.046)} & 0.611{\scriptsize(0.053)} & 0.999{\scriptsize(0.130)}\\
& \textsc{DER} & 
1.483{\scriptsize(0.174)} & 1.180{\scriptsize(0.162)} & 1.450{\scriptsize(0.188)} & 2.480{\scriptsize(0.373)}\\
& \textsc{GRation} & 
\textbf{0.604}{\scriptsize(0.020)} & \textbf{0.557}{\scriptsize(0.037)} & \textbf{0.560}{\scriptsize(0.029)} & \textbf{0.903}{\scriptsize(0.055)} \\
\bottomrule[1.5pt]
\end{tabular}}
\end{minipage} \hfill
  \mbox{}
\end{table*}

\subsubsection{Baselines and Implementations}~Besides the GNN model, we broadly consider baselines from the fields of imbalanced regression and semi-supervised graph learning. Specifically, imbalanced regression baselines include \textsc{LDS}~\cite{yang2021delving}, \textsc{BMSE}~\cite{ren2021bmse}, and \textsc{RankSim}~\cite{gong2022ranksim}. 
The semi-supervised graph learning baseline is \textsc{InfoGraph}~\cite{sun2020infograph} and the graph learning baseline is \textsc{GREA}~\cite{liu2022graph}. To implement \textsc{SGIR} and the baselines, the GNN encoder is GIN~\cite{xu2018how} and the decoder is a three-layer MLP to output property values.
The threshold $\tau$ for selecting confident predictions is determined by the value at a certain percentile of the confidence score distribution. For all the methods, we reports the results on the test sets using the mean (standard deviation) over 10 runs with parameters that are randomly initialized. More Implementation details are in Appendix.

\begin{table*}[htp!]
    \begin{minipage}[c]{0.22\textwidth}
        \centering
         \caption{\small Nine options on the implementation of the label-anchored mixup in~\cref{eq:mix_center_and_real}. Except for the imbalanced labeled graphs $\mathcal{G}_{\text{imb}}$, the additional source of the interval representation $\textbf{z}_i$ and the real graph representation $\textbf{h}_j$ could be $\mathcal{G}_{\text{conf}}$ or $\mathcal{G}_{\text{unlbl}}$. We extensively explore the options for $\mathcal{H}_{\text{aug}}$ and find that source $\textbf{z}_i$ from $\mathcal{G}_{\text{imb}}$ and source $\textbf{h}_j$ from $\mathcal{G}_{\text{unlbl}}$ are usually the best.} \label{table:ablation_aug_exhaustive_comb}
    \end{minipage}\hfill
    \begin{minipage}[c]{0.77\textwidth}
    \centering
    \renewcommand{\arraystretch}{1.1}
    \renewcommand{\tabcolsep}{1.5mm}
    \vspace{-0.4in}
    \resizebox{0.99\linewidth}{!}{\begin{tabular}{@{}cc|cccc|cccc@{}}\toprule[1.2pt]
     \multicolumn{1}{c}{$\textbf{z}_i$} & \multicolumn{1}{c|}{$\textbf{h}_j$} & \multicolumn{4}{c|}{\lipo} & \multicolumn{4}{c}{\oxygen} \\
    \midrule
    \multicolumn{2}{c|}{Additional Source} & All & Many-shot & Med.-shot & Few-shot & All & Many-shot & Med.-shot & Few-shot \\
    \midrule
     None & None
    & 0.439{\scriptsize(0.004)} & 0.361{\scriptsize(0.010)} 
    & 0.419{\scriptsize(0.013)} & 0.529{\scriptsize(0.022)} 
    & 165.5{\scriptsize(12.2)} & 4.7{\scriptsize(1.7)} & 16.5{\scriptsize(7.2)} & 417.4{\scriptsize(31.1)} \\
    None & $\mathcal{G}_{\text{conf}}$
    & 0.447{\scriptsize(0.015)} & 0.359{\scriptsize(0.004)} 
    & 0.423{\scriptsize(0.016)} & 0.549{\scriptsize(0.033)}
    & 158.1{\scriptsize(17.0)} & 4.1{\scriptsize(0.7)} & \textbf{11.3}{\scriptsize(0.7)} & 401.9{\scriptsize(45.1)} \\
    None & $\mathcal{G}_{\text{unlbl}}$ 
    & \textbf{0.432}{\scriptsize(0.012)} & \textbf{0.357}{\scriptsize(0.019)} 
    & \textbf{0.413}{\scriptsize(0.017)} & \textbf{0.515}{\scriptsize(0.020)}
    & \textbf{150.9}{\scriptsize(17.8)} & \textbf{3.8}{\scriptsize(1.1)} & 12.2{\scriptsize(0.6)} & \textbf{382.8}{\scriptsize(46.9)} \\
    \cmidrule{1-6} 
    \cmidrule{7-10} 
    $\mathcal{G}_{\text{conf}}$ & None
    & 0.448{\scriptsize(0.012)} & 0.367{\scriptsize(0.008)} 
    & 0.423{\scriptsize(0.008)} & 0.544{\scriptsize(0.028)}
    & 166.0{\scriptsize(18.2)} & 11.9{\scriptsize(11.3)} & 12.6{\scriptsize(0.9)} & 414.0{\scriptsize(52.6)} \\
    $\mathcal{G}_{\text{conf}}$ & $\mathcal{G}_{\text{conf}}$
    & 0.445{\scriptsize(0.007)} & 0.364{\scriptsize(0.008)} 
    & 0.418{\scriptsize(0.010)} & 0.542{\scriptsize(0.012)}
    & 158.8{\scriptsize(8.4)} & 7.7{\scriptsize(8.9)} & 15.4{\scriptsize(7.8)} & 397.5{\scriptsize(15.4)} \\
    $\mathcal{G}_{\text{conf}}$ & $\mathcal{G}_{\text{unlbl}}$
    & 0.449{\scriptsize(0.021)} & 0.360{\scriptsize(0.023)} 
    & 0.416{\scriptsize(0.016)} & 0.560{\scriptsize(0.039)}
    & 169.5{\scriptsize(56.1)} & 4.5{\scriptsize(1.2)} & 12.7{\scriptsize(1.8)} & 430.4{\scriptsize(145.0)} \\
    \cmidrule{1-6} 
    \cmidrule{7-10} 
    $\mathcal{G}_{\text{unlbl}}$ & None
    & 0.446{\scriptsize(0.007)} & 0.367{\scriptsize(0.009)} 
    & 0.415{\scriptsize(0.011)} & 0.546{\scriptsize(0.011)}
    & 173.1{\scriptsize(30.3)} & 3.7{\scriptsize(0.4)} & 13.5{\scriptsize(1.4)} & 440.0{\scriptsize(79.3)} \\
    $\mathcal{G}_{\text{unlbl}}$ & $\mathcal{G}_{\text{conf}}$
    & 0.446{\scriptsize(0.011)} & 0.368{\scriptsize(0.011)} 
    & 0.421{\scriptsize(0.012)} & 0.539{\scriptsize(0.024)}
    & 174.5{\scriptsize(9.3)} & 8.1{\scriptsize(3.3)} & 11.9{\scriptsize(0.9)} & 440.4{\scriptsize(25.5)} \\
    $\mathcal{G}_{\text{unlbl}}$ & $\mathcal{G}_{\text{unlbl}}$
    & 0.451{\scriptsize(0.007)} & 0.371{\scriptsize(0.012)} 
    & 0.425{\scriptsize(0.008)} & 0.547{\scriptsize(0.015)} 
    & 156.3{\scriptsize(20.5)} & 8.2{\scriptsize(2.9)} & 12.9{\scriptsize(0.9)} & 392.3{\scriptsize(50.6)} \\
    \bottomrule[1.5pt]
    \end{tabular}}
    \vspace{-0.18in}
    \end{minipage}
\end{table*}
\begin{figure*}[ht!]
    \centering
    \subfigure[ \normalsize In the entire label range]{
    \includegraphics[width=0.43\linewidth]{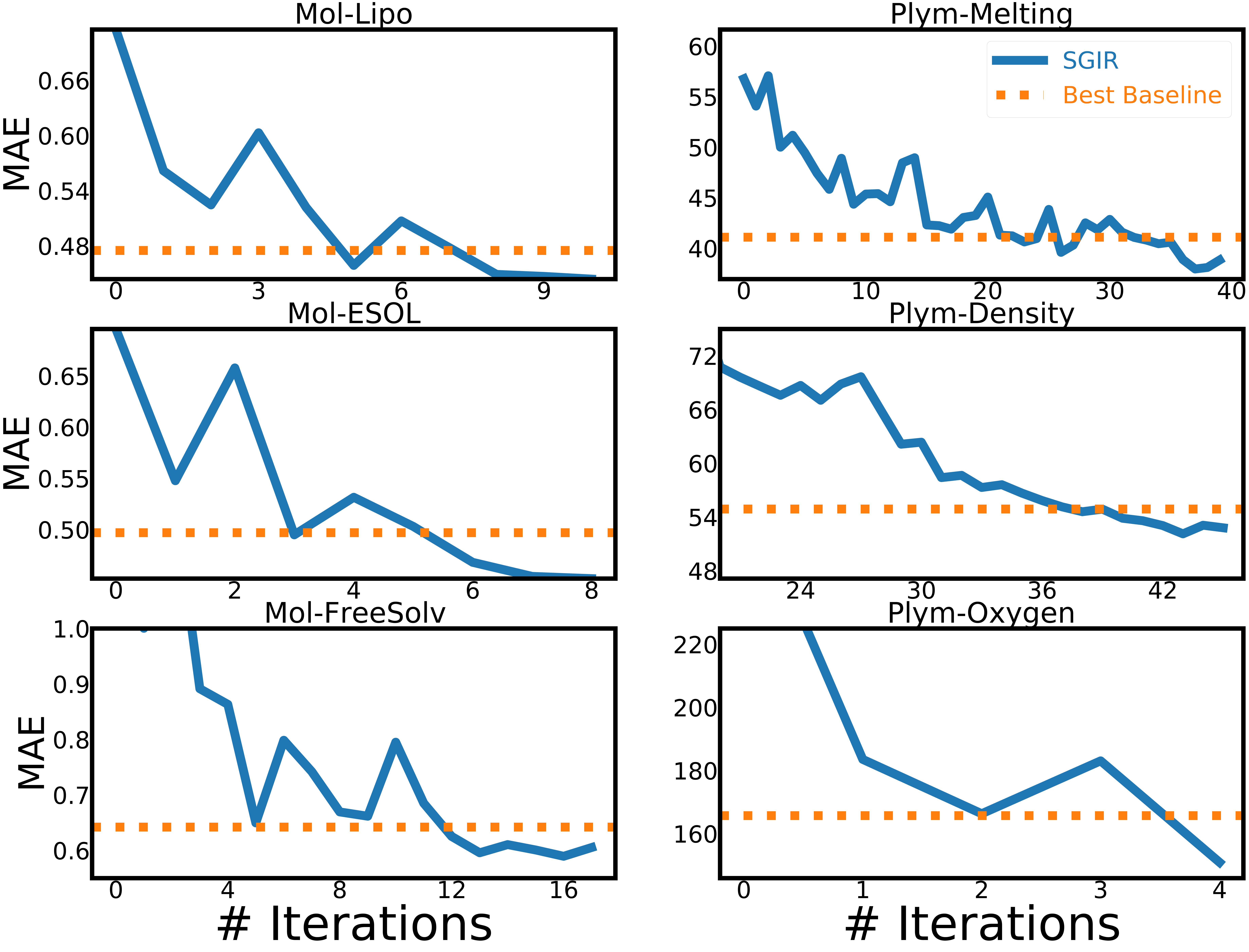}
    \label{fig:iteration_versus_mae_all}
    }
    \hfill
    \subfigure[\normalsize In the \fewrg.]{
    \includegraphics[width=0.43\linewidth]{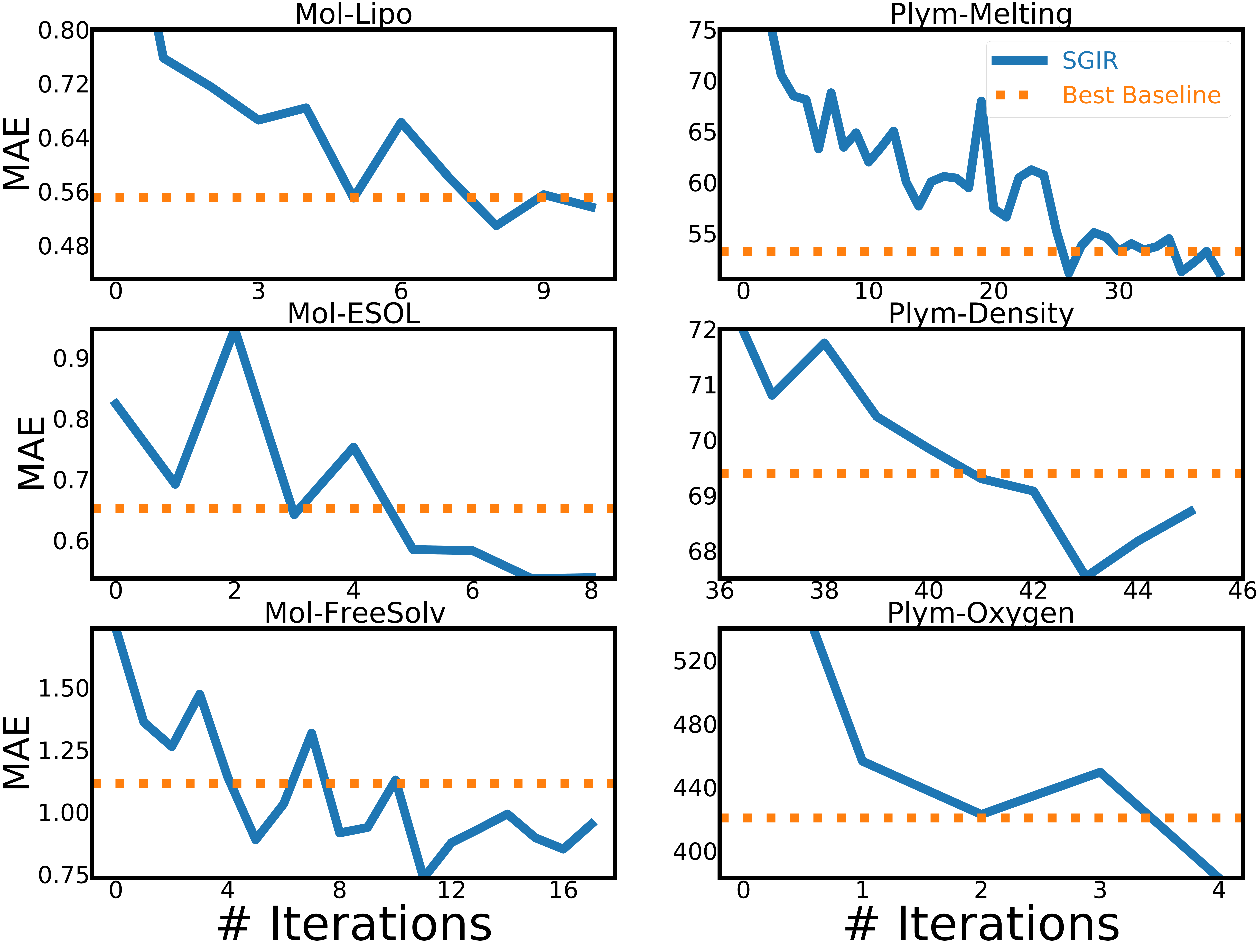}
    \label{fig:iteration_versus_mae_few}
    }
    \vspace{-0.15in}
    \caption{Test performance of \textsc{SGIR} through multiple self-training iterations. MAE for \density is scaled by $\times 1,000$. The iterative self-training algorithm is effective for gradually improving the quality of training data.}
    \label{fig:iteration_versus_mae}
\end{figure*}
\begin{figure*}[t]
    \centering
    \begin{subfigure}
        \centering
        (a) Varying the number $C$ for $\mathcal{G}_\text{conf}$
        \includegraphics[width=0.98\linewidth]{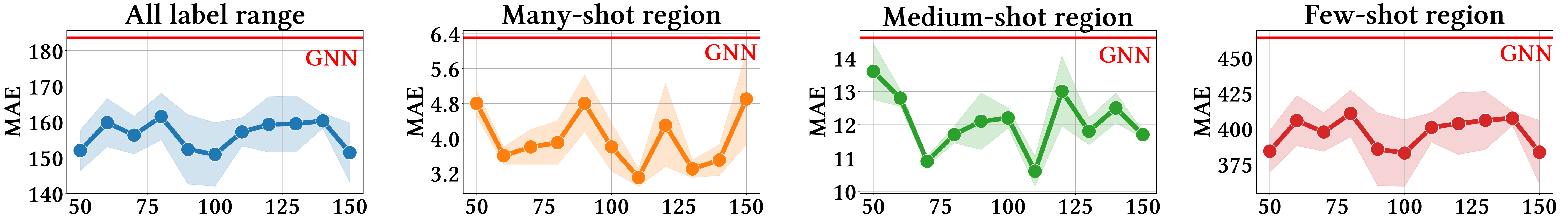} 
        \label{fig:sensi_pseudo}
    \end{subfigure}%
    \hfill
    \begin{subfigure}
        \centering
        (b) Varying the number $C$ for $\mathcal{H}_\text{aug}$
        \includegraphics[width=0.98\linewidth]{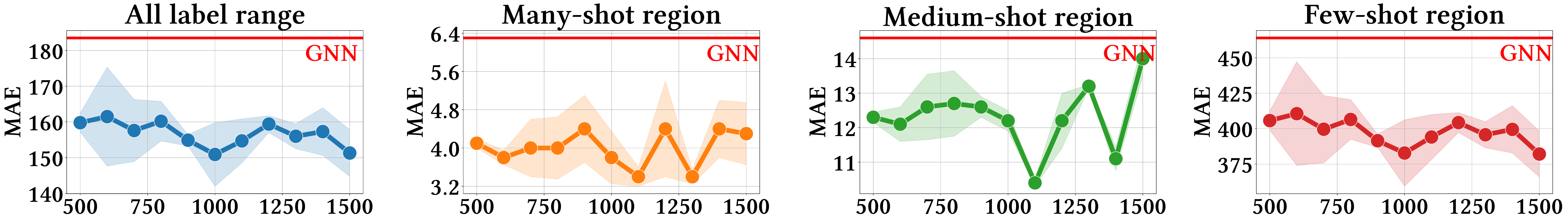}\label{fig:sensi_aug}
    \end{subfigure}%
    \caption{Sensitivity analysis on the number of label intervals ($C$) for pseudo-labeling selection $\mathcal{G}_\text{conf}$ (top) and label-anchored mixup algorithm $\mathcal{H}_\text{aug}$ (bottom). Results are drawn on the Plym-Oxygen.}
    \label{fig:sensi_all_pseudo_aug}
\end{figure*}

\subsection{RQ1: Effectiveness on Property Prediction}
\subsubsection{Effectiveness on Molecule and Polymer Prediction}

\cref{table:six_dataset_results} presents results of \textsc{SGIR} and baseline methods on six graph regression tasks. We have three observations.

\textbf{Overall effectiveness in the entire label range:}
\textsc{SGIR} performs consistently better than competitive baselines on all tasks.
Columns ``All'' in \cref{table:six_dataset_results} show that \textsc{SGIR} reduces MAE over the best baselines (whose MAEs are underlined in the table) relatively by 9.1\%, 8.1\%, and 12.3\% on the three molecule datasets, respectively.
Specifically, on \freesolv, the MAE was reduced from 0.642 to 0.563 with no change on the standard deviation.
This is because \textsc{SGIR} enrich and \emph{balance} the training data with confidently predicted pseudo-labels and augments for data examples on all the possible label ranges, whereas all the baseline models suffer from the bias caused by imbalanced annotations.

\vspace{0.05in}
\textbf{Effectiveness in few-shot label ranges:}
The performance improvements of \method on graph regression tasks are simultaneously from three different label ranges: \manyrg, \mediumrg, and \fewrg.
By looking at the results of baselines, we find that the best performance at a particular range would sacrifice the performance at a different label range.
For example, on the \lipo and \freesolv datasets, while \textsc{GREA} is the second best and best baseline, respectively, in the \manyrg, its performance in the \fewrg is worse than the basic \textsc{GNN} models. Similarly, on the \freesolv dataset, \textsc{LDS} reduces the MAE from \textsc{GNN} relatively by +3.5\% in the \fewrg with a trade-off of a -29\% performance decrease in the \manyrg. Compared to baselines, the improvements from \textsc{SGIR} in the under-represented label ranges are theoretically guaranteed without sacrificing the performance in the well-represented label range. And our experimental observations support the theoretical guarantee, even in more challenging scenarios, \textit{i.e.,} predictions in the label ranges of fewer training shots on smaller datasets.
Specifically, \textsc{SGIR} reduces MAE relatively by 30.3\% and 9.0\% in the \fewrg on \freesolv and \oxygen.
Because \textsc{SGIR} leverages the mutual enhancement of model construction and data balancing: the gradually balanced training data reduce model bias to popular labels; the less biased model improves the quality of pseudo-labels and augmented examples in the \fewrg.

\vspace{0.05in}
\textbf{Effectiveness on different graph regression tasks:}
We observe that the improvements on molecule regression tasks are more significant than those on polymer regression tasks. We hypothesize the reasons to be (1) the quality of unlabeled source data and (2) the size of the label space.
First, our unlabeled graphs consist of more than a hundred thousand unlabeled small molecule graphs from QM9~\cite{ramakrishnan2014quantum} and around ten thousand polymers (macromolecules) from~\cite{liu2022graph}. The massive quantity of unlabeled molecules make it easier to have good quality pseudo-labels and augmented examples for the three small molecule regression tasks on \lipo, \esol, and \freesolv~\cite{ramakrishnan2014quantum}.
Because the majority of unlabeled molecule graphs have a big domain gap with the polymer regression tasks, the quality of expanded training data in polymer regression tasks would be relatively worse than the quality of those in molecule regression. This inspires us to collect more polymer data in the future, even if their properties could not be annotated.
Second, \cref{fig:trainset_dist} has shown that the label ranges in the polymer regression tasks are usually much wider than the ranges in the molecule regression tasks. This poses a great challenge for accurate predictions, especially when we train with a small dataset. 

\subsubsection{Effectiveness on Age Prediction}
Besides molecules and polymers, \cref{table:agedb_results}~presents more results by comparing different methods on the \age dataset. \method consistently improves the model performance compared to the best baselines in different label ranges. In the entire label range, \method reduces the MAE (GM) relatively by +4.7\% (+3.6\%). The advantages mainly stem from the enhancements in the \fewrg, as demonstrated in~\cref{table:agedb_results}, which shows an improvement of +4.3\% and +3.1\% on the MAE and GM metrics, respectively. Different from  \textsc{LDS}, \method improves the model performance for the under-represented and well-represented label ranges at the same time. \cref{table:agedb_results} showcases that the empirical advantages of \method could generalize across different domains.  

\subsection{RQ2: Ablation Studies on Framework Design}
We conduct five studies and analyze the results below. Four ablation studies are (1) $\mathcal{G}_{\text{conf}}$ and $\mathcal{H}_{\text{aug}}$ for data balancing;  (2) mutually enhanced iterative process; (3) choices of confidence score; and (4) quality and diversity of the label-anchored mixup. (5) The sensitivity analysis for the label interval number $C$. Readers can refer to the appendix for complete results.

\subsubsection{Effect of balancing data with different components in $\mathcal{G}_{\text{conf}}$ and $\mathcal{H}_{\text{aug}}$} 
Studies on molecule regression tasks in~\cref{table:ablation} present how \textsc{SGIR} improves the initial supervised performance to the most advanced semi-supervised performance step by step. In the first line for each dataset, we use only imbalanced training data $\mathcal{G}_{\text{conf}}$ to train the regression model and observe that the model performs badly in the \fewrg. The fourth line for each dataset combines the use of regression confidence $\sigma$ and the reverse sampling $p$ to produce $\mathcal{G}_{\text{conf}}$. It improves the MAE performance in the \fewrg relatively by +11.2\%, +3.2\%, and +15.9\% on the \lipo, \esol, and \freesolv datasets, respectively. 
The label-anchored mixup algorithm produces the augmented graph representations $\mathcal{H}_{\text{aug}}$ for the under-represented label ranges. By applying $\mathcal{H}_{\text{aug}}$ with $\mathcal{G}_{\text{conf}}$, the last line continues improving the MAE performance in the \fewrg (compared to the third line) relatively by +3.3\%, +1.3\%, and +6.5\% on the \lipo, \esol, and \freesolv datasets, respectively. Because the use of $\mathcal{H}_{\text{aug}}$ provides a chance to lead the label distributions of training data closer to a perfect balance.  Specifically, the effect of semi-supervised pseudo-labeling, or $\mathcal{G}_{\text{conf}}$, comes from the regression confidence $\sigma$ and reverse sampling rate $p$. Results on \esol and \freesolv show that without the confidence $\sigma$ (the second line), reverse sampling was useless due to heavy label noise. Results on all molecule datasets indicate that without the reverse sampling rate $p$ (the first line), the improvement to \fewrg by pseudo-labels was limited.

\subsubsection{Effect of iterative self-training}
\cref{fig:iteration_versus_mae} confirms that model learning and balanced training data mutually enhance each other in \textsc{SGIR}. Because we find that the model performance gradually approximates and outperforms the best baseline in the entire label range, as well as the \fewrg, after multiple iterations. It also indicates that the quality of the training data is steadily improved over iterations. Especially for the under-represented label ranges.

\subsubsection{Effect of regression confidence measurements} \cref{table:uncertainty} shows that compared to existing methods that could define regression confidence, the measurement we define and use, \textsc{GRation}, is the best option for evaluating the quality of pseudo-labels in graph regression tasks. Because \textsc{GRation} uses various environments subgraphs, which provide diverse perturbations for robust graph learning~\cite{liu2022graph}. We also observe that \textsc{Dropout} can be a good alternative of \textsc{GRation}. \textsc{Dropout} has extensive assessments~\cite{gal2016dropout} and makes it possible for \textsc{SGIR} to be extended to regression tasks for other data types such as images and texts.

\subsubsection{Effect of label-anchored mixup augmentation}
We implement $\mathbf{z}_i$ using $\mathcal{G}_{\text{imb}}$ to improve the augmentation quality and $\mathcal{G}_{\text{imb}}\cup\mathcal{G}_{\text{unlbl}}$ to improve the diversity. \cref{table:ablation_aug_exhaustive_comb} presents extensive empirical studies to support our idea. It shows that when many noisy representation vectors from unlabeled graphs are included in the interval center $\mathbf{z}_i$, the quality of augmented examples is relatively low, which degrades the model performance in different label ranges. On the other hand, the representations of unlabeled graphs improve the diversity of the augmented examples when we assign low mixup weights to them as in~\cref{eq:mix_center_and_real}. Considering both quality and diversity, the effectiveness of the algorithm is further demonstrated in \cref{table:ablation} by significantly reducing the errors for rare labels. From the fifth line of each dataset in~\cref{table:ablation}, we find that it is also promising to directly use the label-anchored mixup augmentation (as $\mathcal{G}_{\text{imb}}\cup\mathcal{H}_{\text{aug}}$) for data balancing. Although its performance may be inferior to the performance using $\mathcal{G}_{\text{imb}}\cup\mathcal{G}_{\text{conf}}$ (as the third line of each dataset in~\cref{table:ablation}), the potential of the label-anchored mixup algorithm could be further enhanced by improving the quality of the augmented examples to close the gap with real molecular graphs.

\subsubsection{Sensitivity analysis for the label interval number $C$}
We find the best values of $C$ in main experiments using the validation set for pseudo-labeling and label-anchored mixup. We suggest setting the number $C$ to approximately 100 for pseudo-labeling and around 1,000 for label-anchored mixup. Specifically, sensitivity analysis is conducted on the Plym-Oxygen dataset to analyze the effect of the number $C$. Results are presented in~\cref{fig:sensi_all_pseudo_aug}. We observe that \method is robust to a wide range of choices for the number of intervals.

\section{Conclusions}\label{sec:conclusion}
In this work, we explored a novel graph imbalanced regression task and improved semi-supervised learning on it. We proposed a self-training framework to gradually reduce the model bias of data imbalance through multiple iterations. In each iteration, we selected more high-quality pseudo-labels for rare label values and continued augmenting training data to approximate the perfectly balanced label distribution. Experiments demonstrated the effectiveness and reasonable design of the proposed framework, especially on material science.

\begin{acks}
This work was supported in part by NSF IIS-2142827, IIS-2146761, and ONR N00014-22-1-2507.
\end{acks}

\bibliographystyle{ACM-Reference-Format}
\bibliography{ref}

\begin{table*}[ht]
\caption{Comparing SGIR with related methods on research problem settings.}
\label{tab:methods}
\centering
\resizebox{0.7\linewidth}{!}{
\begin{tabular}{l|cccc}
\toprule
& Is \textbf{S}emi-supervised & Learning & Addressing & Solving \\
& method? & \textbf{G}raph data? & \textbf{I}mbalance? & \textbf{R}egression?  \\
(Otherwise, assuming:) & {\small (Supervised)} & {\small (Non-graph)} & {\small (Balance) } & {\small (Classification)} \\
\midrule
\darp~\cite{kim2020distribution} & \cmark  &  & \cmark & \\
\daso~\cite{oh2022distribution} & \cmark  &  & \cmark & \\
\bisampling~\cite{he2021rethinking} & \cmark  &  & \cmark & \\
\cadr~\cite{hu2022on} & \cmark  &  & \cmark & \\
\crest~\cite{wei2021crest} & \cmark  &  & \cmark & \\
\midrule
\lds~\cite{yang2021delving} &  &  & \cmark & \cmark \\
\bmse~\cite{ren2021bmse} &  &  & \cmark & \cmark   \\
\textsc{RankSim}~\cite{gong2022ranksim} &  &  & \cmark & \cmark   \\
\midrule
\ssdkl~\cite{jean2018semi} & \cmark &  &  &\cmark  \\
\infograph ~\cite{sun2020infograph} & \cmark  & \cmark & & \cmark \\
\midrule
\method (Ours) & \cmark & \cmark & \cmark & \cmark\\
\bottomrule
\end{tabular}
}
\end{table*}
\newpage
\appendix
\section{Further Related Work}

\subsection{A Systematic Comparison with Related Methods}
We compare \textsc{SGIR} with a line of related work on four important settings of research problem in~\cref{tab:methods}. From the table we find that existing work mostly focused on solving imbalance problems in semi-supervised classification tasks with categorical labels and non-graph data. There lacks an exploration of research on semi-supervised learning and imbalance learning for graph regression.

\subsection{Sampling Strategy in Self-training}
To the best of our knowledge, reverse sampling is one of the most suitable sampling strategies to address class imbalance issues in self-training~\cite{wei2021crest}. Compared to other strategies like random sampling or mean sampling~\cite{he2021rethinking}, reverse sampling is also the most suitable one for graph imbalanced regression. This is because reverse sampling compensates for the label imbalance and enriches training examples. Other strategies cannot make the training data more balanced. They would lead the prediction model to be still biased to the majority of data. More complex sampling strategies that combine reverse sampling, mean sampling, and random sampling would be a promising direction for future work.

\section{Proofs of Theoretical Motivations}\label{add:sec:proof}
We rely on two theorems to derive~\cref{theo:motivation}.
\subsection{Existing Theorems}
Given a classifier $f$ from the function class $\mathcal{F}$, an input example $x$ from the feature space $\mathcal{X}$ and its label $y$.
\begin{theorem}[from \cite{bartlett2002rademacher,kakade2008complexity}]\label{add:theorem: exising 1}
    Assume the expected loss on examples is $\mathcal{E}[f]$ and the corresponding empirical loss $\hat{\mathcal{E}[f]}$. Assume the loss is Lipschitz with Lipschitz constant $L_e$. And it is bounded by $c_0$. For any $\delta > 0$ and with probability at least $1-\delta$ simultaneously for all $f \in \mathcal{F}$ we have that
    \begin{equation}
    \mathcal{E}[f] \leq \hat{\mathcal{E}}[f]+2 L_e\mathcal{R}_n(\mathcal{F})+c_0\sqrt{\frac{\log (1 / \delta)}{2 n}},
    \end{equation}
    where $n$ is the number of example and $ \hat{\mathcal{R}}_n (\mathcal{F}) $ is the Rademacher complexity measurement of the hypothesis class $\mathcal{F}$.
\end{theorem}

\begin{theorem}[from \cite{kakade2008complexity}]\label{add:theorem: exising 2}
Applying \cref{add:theorem: exising 1} and considering the fraction of data having $\gamma$-margin mistakes, or $K_\gamma [f]:= \frac{|i: y_i f(x_i) < \gamma |}{n}$. Assume $\forall f \in \mathcal{F}$ we have $\operatorname{sup}_{x \in \mathcal{X}} |f(x)| \leq c_1$. Then, with probability at least $1-\delta$ over the example, for all margins $\gamma >0$ and all $f \in \mathcal{F}$ we have, 
    \begin{align}
        \mathcal{E}[f] & \leq K_\gamma[f]+4\frac{\mathcal{R}_n(\mathcal{F})}{\gamma}+\sqrt{\frac{2 \log \left(\log _2(4 c_1 / \gamma)\right) + \log (1 / \delta)}{2 n}}, \\
        & \leq K_\gamma[f]+4 \frac{\mathcal{R}_n(\mathcal{F})}{\gamma}+\sqrt{\frac{\log \left(\log _2 \frac{4 c_1}{\gamma}\right)}{n}}+\sqrt{\frac{\log (1 / \delta)}{2 n}}.
    \end{align}
\end{theorem}

\subsection{Proof of theorem 4.1}

In our work, we use the regression function $f$ to predict the label value. We calculate the reciprocal of the distance between the predicted label and interval centers as unnormalized probabilities of the graph $S_{[b_i, b_{i+1})}(G)$ being assigned to the interval $[b_i, b_{i+1}), i \in \{1, 2, \dots, C\}$.
Given a hard margin $\gamma$, we use $\mathcal{E}_{\gamma, [b_i, b_{i+1})}[f]$ to denote the hard margin loss for examples in the interval $[b_i, b_{i+1})$:
\begin{equation}
   \mathcal{E}_{\gamma, [b_i, b_{i+1})}[f] =  \underset{G \sim \mathcal{P}_{[b_i, b_{i+1})}}{\operatorname{Pr}} \left[S_{[b_i, b_{i+1})}(G) < \max_{j \neq i}  S_{[b_j, b_{j+1})}(G) + \gamma \right].
\end{equation}
We assume its empirical variant is $\hat{\mathcal{E}}_{\gamma, [b_i, b_{i+1})}[f]$. The empirical Rademacher complexity $ \hat{\mathcal{R}}_{(b_i, b_{i+1}]} (\mathcal{F}) $ is used as the complexity measurement $\textup{C}(\mathcal{F})$ for the hypothesis class $\mathcal{F}$. With a vector $\sigma$ of i.i.d. uniform $\{-1, +1\}$ bits, we have
\begin{align}
    & \hat{\mathcal{R}}_{(b_i, b_{i+1}]}(\mathcal{F}) = \\ & \frac{1}{n_{(b_i, b_{i+1}]}} \mathbb{E}_\sigma\left[\sup _{f \in \mathcal{F}} \sum_{G_i \in [b_i, b_{i+1})} \sigma_i\left[S_{[b_i, b_{i+1})}\left(G_i\right) -\max _{j \neq i} S_{[b_j, b_{j+1})}\left(G_i\right)\right]\right]
\end{align}
As any $G_i$ in the interval ${(b_i, b_{i+1}]}$ is an i.i.d. sample from the distribution $\mathcal{P}_{[b_i, b_{i+1})}$, we directly apply the standard margin-based generalization bound \cref{add:theorem: exising 2}~\cite{kakade2008complexity}: with probability $1-\delta$, for all choices of $\gamma_{_{[b_i, b_{i+1})}} > 0$ and $f \in \mathcal{F}$,
\begin{align}
    \mathcal{E}_{[b_i, b_{i+1})} & \leq 
    \hat{\mathcal{E}}_{\gamma, [b_i, b_{i+1})}[f]+
    4\frac{\hat{\mathcal{R}}_{(b_i, b_{i+1}]}(\mathcal{F})}{\gamma_{_{[b_i, b_{i+1})}}} \label{add:eq:replace} \\ & \quad\quad\quad\quad +
    \sqrt{\frac{2 \log \left(\log _2( \frac{4 c_1}{\gamma_{_{[b_i, b_{i+1})}}})\right) + \log (1 / \delta)}{2 n_{[b_i, b_{i+1})}}}, \nonumber \\
    &  \leq  \hat{\mathcal{E}}_{\gamma, [b_i, b_{i+1})}[f] + \frac{1}{\gamma_{_{[b_i, b_{i+1})}}}  \sqrt{\frac{\textup{C}(\mathcal{F}) }{n_{[b_i, b_{i+1})}}} \label{add:eq:complexity} \\ & \quad\quad\quad\quad +  \sqrt{\frac{2 \log \left(\log _2( \frac{4 c_1}{\gamma_{_{[b_i, b_{i+1})}}})\right)  \log (1 / \delta)}{2 n_{[b_i, b_{i+1})}}}, \nonumber \\
    & \lessapprox  \frac{1}{\gamma_{_{[b_i, b_{i+1})}}}  \sqrt{\frac{\textup{C}(\mathcal{F}) }{n_{[b_i, b_{i+1})}}} +  \sqrt{\frac{\log\log_2(1/\gamma_{_{[b_i, b_{i+1})}}) +\log (1 / \delta)}{{2 n_{[b_i, b_{i+1})}}}} \label{add:eq:approximate}.
\end{align}
We derive \cref{add:eq:complexity} from \cref{add:eq:replace} because the Rademacher complexity $\hat{\mathcal{R}}_{(b_i, b_{i+1}]}(\mathcal{F})$ typically scales as $\sqrt{\frac{\textup{C}(\mathcal{F})}{n_{(b_i, b_{i+1}]}}}$ for some complexity measurement $\textup{C}(\mathcal{F})$~\cite{cao2019learning}. We derive \cref{add:eq:approximate} from \cref{add:eq:complexity} by ignoring constant factors~\cite{cao2019learning}. Since the overall performance $\mathcal{E}_{\textup{bal}}[f]$ is calculated over all intervals, we get it as $\mathcal{E}_{\textup{bal}}[f] = \frac{1}{C}\sum_{i=1}^C \mathcal{E}_{[b_i, b_{i+1})}$. 

\subsection{Discussions}\label{add:sec:theorem_discussion}

Existing work on the theoretical analysis of mixup~\cite{liu2023overtraining,carratino2020mixup,zhang2021how} mainly focused on image classification and cannot guarantee that the augmented graph examples are i.i.d sampled from the conditional distribution $\mathcal{P}_{[b_i, b_{i+1}]}$ for a specific interval $[b_i, b_{i+1}]$. While a recent work proposed the C-mixup~\cite{yao2022c} to sample closer pairs of examples with higher probability for regression tasks, it did not fit our theoretical motivation to address the label imbalance issue: with C-mixup, the pairs in the over-represented label ranges have a higher probability to be sampled than the under-represented ones. Compared to these theories and methods for the mixup algorithm, our label-anchored mixup allows direct application to imbalanced regression tasks without compromising the assumption in our theoretical motivation. This is because we use the augmented virtual examples $\mathcal{H}_\text{aug}$ based on the label anchor within intervals $[b_i, b_{i+1}]$. Augmented examples are independently created with~\cref{eq:mix_center_and_real}. Since the interval centers could be mixed with any other real graphs from $\mathcal{G}_\text{imb} \cup \mathcal{G}_\text{conf}$, any value in the interval space could be sampled. Besides, it is reasonable to use the distribution of the entire label space (from $\mathcal{G}_\text{imb} \cup \mathcal{G}_\text{conf}$) to approximate the distribution within the interval and assume that the conditional distribution $\mathcal{P}_{[b_i, b_{i+1}]}$ does not change.

We build the theoretical principle for imbalanced regression with intervals to connect with existing theoretical principles for classification. Future theoretical work on imbalanced regression can leverage the advantages of using mixture regressor models~\cite{sugiyama2006mixture}, which have been used to address covariate shift problems in regression tasks. Additionally, exploring the promising connection between domain adaptation theories and sample selection bias~\cite{sun2016return,cortes2008sample} holds potential for further advancements in this field.

\section{Experiments}
\subsection{Dataset Details}
We give a comprehensive introduction to our datasets used for regression tasks and splitting idea from~\cite{yang2021delving, gong2022ranksim}.

\paragraph{\lipo} It is a dataset to predict the property of lipophilicity consisting of 4200 molecules. The lipophilicity is important for solubility and membrane permeability in drug molecules. This dataset originates from ChEMBL~\cite{mendez2019chembl}. The property is from experimental results for the octanol/water distribution coefficient ($\log D$ at pH 7.4). 

\paragraph{\esol} It is to predict the water solubility ($\log$ solubility in mols per litre) from chemical structures consisting of 1128 small organic molecules.

\paragraph{\freesolv} It is to predict the hydration free energy of molecules in water consisting of 642 molecules. The property is experimentally measured or calculated.

\paragraph{\meltTemp} It is used to predict the property of melting temperature ($^\circ$C). It is collected from PolyInfo, a web-based polymer database~\cite{otsuka2011polyinfo}.

\paragraph{\density} It is used to predict the property of polymer density (g/cm$^3$). It is collected from PolyInfo, a web-based polymer database~\cite{otsuka2011polyinfo}.

\paragraph{\oxygen} It is used to predict the property of oxygen permeability (Barrer). It is created from the Membrane Society of Australasia portal consisting of experimentally measured gas permeability data \cite{thornton2012polymer}.

\paragraph{Unlabeled Data for Molecules and Polymers} The total number of unlabeled graphs for molecule and polymers is 146,129, consisting of 133,015 molecules from QM9~\cite{ramakrishnan2014quantum} and 13,114 monomers (the repeated units of polymers) from~\cite{liu2022graph}. QM9 is a molecule dataset for stable small organic molecules consisting of atoms C, H, O, N, and F.  We use it as a source of unlabeled data. 
We integrate four polymer regression datasets including \meltTemp, \density, \oxygen and another one from~\cite{liu2022graph} for the glass transition temperature as the other source of unlabeled data. 
We note that the unlabeled graphs may be slightly less than 146,129 for a polymer task on \meltTemp, \density or \oxygen. It is because we remove the overlapping graphs for the current polymer task with the polymer unlabeled data.

\paragraph{Data splitting for Molecules and Polymers} We split the datasets based on the approach in previous works~\cite{yang2021delving, gong2022ranksim} motivated for two reasons. First, we want the training sets to well characterize the imbalanced label distribution as presented in the original datasets. Second, we want relatively balanced valid and test sets to fairly evaluate the model performance in different ranges of label values. 

\paragraph{\age}
The details of the age regression dataset are presented in~\cref{tab:dataset_stat} (\age) and~\cref{fig:age_distribution}. The graph dataset \age is constructed from image superpixels using the algorithms from~\cite{knyazev2019understanding} on the image dataset \emph{AgeDB-DIR} from~\cite{moschoglou2017agedb, yang2021delving}. Each face image in \emph{AgeDB-DIR} has an age label from 0 to 101. We fisrt compute the SLIC superpixels for each image without losing the label-specific information~\cite{achanta2012slic,knyazev2019understanding}. Then we use the superpixels as nodes and calculate the spatial distance between superpixels to build edges for each image~\cite{knyazev2019understanding}. Binary edges are constructed between superpixel nodes by applying a threshold on the top-5\% of the smallest spatial distances. After building a graph for each image, we follow the data splitting in~\cite{yang2021delving} to study the imbalanced regression problem. We randomly remove 70\% labels in the training/validation/test data and use them as unlabeled graphs. Finally, the graph dataset \age consists of 3,619 graphs for training, 628 graphs for validation, 628 graphs for testing, and 11,613 unlabeled graphs for semi-supervised learning.

\subsection{Implementation Details} We use the Graph Isomorphism Network (GIN)~\cite{xu2018how} as the GNN encoder for $f_\theta$ to get the graph representation and three layers of Multilayer perceptron (MLP) as the decoder to predict graph properties. 
The threshold $\tau$ for selecting confident predictions is determined by the value at a certain percentile of the confidence score distribution. To implement it, we set it up as a hyperparameter $\tau_\text{pct}$ determining the percentile value of the prediction variance (\textit{i.e.,} the reciprocal of confidence) of the labeled training data.
In experiments, all methods are implemented on Linux with Intel Xeon Gold 6130 Processor (16 Cores @2.1Ghz), 96 GB of RAM, and a RTX 2080Ti card (11 GB RAM). For all the methods, we reports the results on the test sets using the mean (standard deviation) over 10 runs with parameters that are randomly initialized. Note that the underlying design of the graph learning model used in \textsc{SGIR} is \textsc{GREA} with a learning objective as follows.
Given $(G, y) \in \mathcal{G}_{\text{imb}}\cup \mathcal{G}_{\text{conf}} $, \textsc{GREA}~\cite{liu2022graph} will output a vector $\mathbf{m} \in \mathbb{R}^K$ that indicates the probability of $K$ nodes in a graph being in the rationale subgraph.
So, we could get $\mathbf{h}^{(r)} = \mathbf{1}^{\top}_K \cdot (\mathbf{m} \times \mathbf{H}) $ and $\mathbf{h}^{(e)} = \mathbf{1}^{\top}_K \cdot ((\mathbf{1}_K - \mathbf{m}) \times \mathbf{H})$, where $\mathbf{H} \in \mathbb{R}^{K \times d}$ is the node representation matrix. By this, the optimization objectives of a graph consist of 
\begin{equation}\label{eq:appendix_top_rationalization}
    \begin{cases}
      & \ell_\text{imb+conf} = \operatorname{MAE}(f(\mathbf{h}^{(r)}), y) + \mathbb{E}_{G^\prime} \big[ \operatorname{MAE}(f(\mathbf{h} + \mathbf{h}^\prime  ), y) \big]  \\
      & \quad \quad \quad + \operatorname{Var}_{G^\prime} \big( \{ \operatorname{MAE}(f(\mathbf{h} + \mathbf{h}^\prime ), y) \} \big), \\
      \\
      & \ell_\text{regu} = \frac{1}{K} \sum_{k=1}^K  |\mathbf{m}_k| - \gamma 
    \end{cases}   \nonumber   
\end{equation}
$\ell_\text{regu}$ regularizes the vector $\mathbf{m}$ and $\gamma \in [0,1]$ is a hyperparameter to control the expected size of $G^{(r)}$. $G^\prime$ is the possible graph in the same batch that provides environment subgraphs and $\mathbf{h}^\prime$ is the representation vector of the environment subgraph.
When combining the rationale-environment pairs to create new graph examples, the original GREA creates the same number of examples for the under-represented rationale and the well/over-represented rationale. We observe that it may make the training examples more imbalanced. Therefore, we use the reweighting technique to penalize more for the expectation term ($\mathbb{E}_{G^\prime} \big[ \operatorname{MAE}(f(\mathbf{h} + \mathbf{h}^\prime  ), y) \big]$) and variance term ($\operatorname{Var}_{G^\prime} \big( \{ \operatorname{MAE}(f(\mathbf{h} + \mathbf{h}^\prime ), y) \} \big)$) in $\ell_\text{imb+conf}$ when the label is from the under-represented ranges. The weight of the expectation and variance terms for a graph with label $y$ is
\begin{equation}\label{eq:opt_exp_var_reformulated}
    w = \frac{ \exp (\sum_{b=1}^B |y - y_b |/t )}{ \exp( \sum_{j=1}^B \sum_{b=1}^B |y - y_b | / t )}, \nonumber  
\end{equation}
where $B$ is the batch size and $t$ is the temperature hyper-parameter.

\subsection{Additional Experimental Results}
\begin{table*}[ht]
\centering
\caption{Complete results of ablation study and mixup options ({MAE $\downarrow$} and GM $\downarrow$) on three molecule datasets. The best mean is \textbf{bolded}. For the label-anchored mixup options, the first column is the source of $\textbf{z}_i$ and the second column is the source of $\textbf{h}_j$.}
\label{table:ablation_all_mol}
\setlength{\tabcolsep}{2pt}
\resizebox{0.98\linewidth}{!}{\begin{tabular}{@{}c|ll|ccccccccc@{}}\toprule[1.2pt]
\multicolumn{3}{c}{} & \multicolumn{4}{c}{{MAE $\downarrow$}} & \phantom{abc}& \multicolumn{4}{c}{{GM $\downarrow$}} \\
\cmidrule{4-7} \cmidrule{9-12} 
\multicolumn{3}{c}{} & All & Many-shot & Med.-shot & Few-shot &&  All & Many-shot & Med.-shot & Few-shot \\

\midrule[1.2pt]
\multicolumn{12}{c}{\lipo}  \\
\midrule
\parbox[t]{5mm}{\multirow{5}{*}{\rotatebox[origin=c]{90}{Ablation Study}}} 
& \multicolumn{2}{l|}{$\mathcal{G}_{\text{imb}}$}
& 0.477{\scriptsize(0.014)} & 0.378{\scriptsize(0.030)} 
& 0.440{\scriptsize(0.011)} & 0.600{\scriptsize(0.006)} &
& 0.288{\scriptsize(0.008)} & 0.236{\scriptsize(0.015)} 
& 0.267{\scriptsize(0.013)} & 0.371{\scriptsize(0.017)} \\
& \multicolumn{2}{l|}{ $\mathcal{G}_{\text{imb}}\cup\mathcal{G}_{\text{conf}}$ }
& 0.442{\scriptsize(0.012)} & 0.372{\scriptsize(0.007)} 
& 0.415{\scriptsize(0.004)} & 0.533{\scriptsize(0.026)} &
& 0.267{\scriptsize(0.013)} & 0.240{\scriptsize(0.008)} 
& 0.245{\scriptsize(0.016)} & 0.320{\scriptsize(0.027)} \\
& \multicolumn{2}{l|}{ $\mathcal{G}_{\text{imb}}\cup\mathcal{G}_{\text{conf}}$ (w/o  $\sigma$)}
& 0.446{\scriptsize(0.008)} & \textbf{0.356}{\scriptsize(0.003)} 
& \textbf{0.407}{\scriptsize(0.011)} & 0.564{\scriptsize(0.016)} &
& 0.272{\scriptsize(0.006)} & \textbf{0.222}{\scriptsize(0.002)} 
& \textbf{0.244}{\scriptsize(0.008)} & 0.363{\scriptsize(0.013)} \\
& \multicolumn{2}{l|}{ $\mathcal{G}_{\text{imb}}\cup\mathcal{G}_{\text{conf}}$ (w/o  $p$)}
& 0.448{\scriptsize(0.006)} & 0.371{\scriptsize(0.004)} 
& 0.421{\scriptsize(0.012)} & 0.543{\scriptsize(0.016)} &
& 0.270{\scriptsize(0.002)} & 0.228{\scriptsize(0.009)} 
& 0.255{\scriptsize(0.008)} & 0.333{\scriptsize(0.015)} \\
& \multicolumn{2}{l|}{  $\mathcal{G}_{\text{imb}}\cup\mathcal{H}_{\text{aug}}$ }
& 0.456{\scriptsize(0.007)} & 0.372{\scriptsize(0.014)} & 0.436{\scriptsize(0.010)} & 0.549{\scriptsize(0.005)} &
& 0.278{\scriptsize(0.013)} & 0.235{\scriptsize(0.019)} & 0.265{\scriptsize(0.014)} & 0.338{\scriptsize(0.006)}  \\
& \multicolumn{2}{l|}{  $\mathcal{G}_{\text{imb}}\cup\mathcal{G}_{\text{conf}}\cup \mathcal{H}_{\text{aug}}$ }
& \textbf{0.432}{\scriptsize(0.012)} & 0.357{\scriptsize(0.019)} & 0.413{\scriptsize(0.017)} & \textbf{0.515}{\scriptsize(0.020)} && \textbf{0.264}{\scriptsize(0.013)} & 0.224\scriptsize(0.016) & 0.256{\scriptsize(0.017)} & \textbf{0.314}{\scriptsize(0.015)} \\

%%%%%%%%%%% D_a
\cmidrule{1-7} \cmidrule{9-12} 
\parbox[t]{5mm}{\multirow{9}{*}{\rotatebox[origin=c]{90}{$\textbf{z}_i$ and $\textbf{h}_j$ options in Mixup}}}
&\multirow{3}{*}{\shortstack{\\  $\mathcal{G}_{\text{imb}}$ } }
& $\mathcal{G}_{\text{imb}}$
& 0.439{\scriptsize(0.004)} & 0.361{\scriptsize(0.010)} 
& 0.419{\scriptsize(0.013)} & 0.529{\scriptsize(0.022)} &
& 0.267{\scriptsize(0.005)} & 0.231{\scriptsize(0.015)} 
& 0.256{\scriptsize(0.010)} & 0.318{\scriptsize(0.020)} \\
&& $\mathcal{G}_{\text{imb}} \cup \mathcal{G}_{\text{conf}} $
& 0.447{\scriptsize(0.015)} & 0.359{\scriptsize(0.004)} 
& 0.423{\scriptsize(0.016)} & 0.549{\scriptsize(0.033)} &
& 0.274{\scriptsize(0.017)} & \textbf{0.221}{\scriptsize(0.007)} 
& 0.264{\scriptsize(0.020)} & 0.344{\scriptsize(0.031)} \\
&& $\mathcal{G}_{\text{imb}} \cup \mathcal{G}_{\text{unlbl}} $
& \textbf{0.432}{\scriptsize(0.012)} & \textbf{0.357}{\scriptsize(0.019)} 
& \textbf{0.413}{\scriptsize(0.017)} & \textbf{0.515}{\scriptsize(0.020)} &
& \textbf{0.264}{\scriptsize(0.013)} & 0.224{\scriptsize(0.016)} 
& 0.256{\scriptsize(0.017)} & \textbf{0.314}{\scriptsize(0.015)} \\
\cmidrule{2-7} \cmidrule{9-12} 
&\multirow{3}{*}{\shortstack{\\  $\mathcal{G}_{\text{imb}} \cup \mathcal{G}_{\text{conf}} $}  }
& $\mathcal{G}_{\text{imb}}$
& 0.448{\scriptsize(0.012)} & 0.367{\scriptsize(0.008)} 
& 0.423{\scriptsize(0.008)} & 0.544{\scriptsize(0.028)} &
& 0.270{\scriptsize(0.013)} & 0.230{\scriptsize(0.013)} 
& 0.257{\scriptsize(0.014)} & 0.328{\scriptsize(0.025)}\\
&& $\mathcal{G}_{\text{imb}} \cup \mathcal{G}_{\text{conf}} $
& 0.445{\scriptsize(0.007)} & 0.364{\scriptsize(0.008)} 
& 0.418{\scriptsize(0.010)} & 0.542{\scriptsize(0.012)} &
& 0.271{\scriptsize(0.009)} & 0.227{\scriptsize(0.011)} 
& 0.256{\scriptsize(0.011)} & 0.337{\scriptsize(0.016)} \\
&& $\mathcal{G}_{\text{imb}} \cup \mathcal{G}_{\text{unlbl}} $
& 0.449{\scriptsize(0.021)} & 0.360{\scriptsize(0.023)} 
& 0.416{\scriptsize(0.016)} & 0.560{\scriptsize(0.039)} &
& 0.270{\scriptsize(0.019)} & 0.223{\scriptsize(0.017)} 
& 0.255{\scriptsize(0.019)} & 0.340{\scriptsize(0.032)} \\
\cmidrule{2-7}  \cmidrule{9-12} 
&\multirow{3}{*}{\shortstack{\\  $\mathcal{G}_{\text{imb}} \cup \mathcal{G}_{\text{unlbl}} $}  }
& $\mathcal{G}_{\text{imb}}$
& 0.446{\scriptsize(0.007)} & 0.367{\scriptsize(0.009)} 
& 0.415{\scriptsize(0.011)} & 0.546{\scriptsize(0.011)} &
& 0.268{\scriptsize(0.006)} & 0.228{\scriptsize(0.008)} 
& \textbf{0.248}{\scriptsize(0.005)} & 0.336{\scriptsize(0.012)} \\
&& $\mathcal{G}_{\text{imb}} \cup \mathcal{G}_{\text{conf}} $
& 0.446{\scriptsize(0.011)} & 0.368{\scriptsize(0.011)} 
& 0.421{\scriptsize(0.012)} & 0.539{\scriptsize(0.024)} &
& 0.270{\scriptsize(0.004)} & 0.233{\scriptsize(0.010)} 
& 0.249{\scriptsize(0.009)} & 0.334{\scriptsize(0.017)} \\
&& $\mathcal{G}_{\text{imb}} \cup \mathcal{G}_{\text{unlbl}} $
& 0.451{\scriptsize(0.007)} & 0.371{\scriptsize(0.012)} 
& 0.425{\scriptsize(0.008)} & 0.547{\scriptsize(0.015)} &
& 0.273{\scriptsize(0.008)} & 0.222{\scriptsize(0.007)} 
& 0.260{\scriptsize(0.012)} & 0.344{\scriptsize(0.014)}  \\

\midrule
\multicolumn{12}{c}{\esol}  \\
\midrule
\parbox[t]{5mm}{\multirow{5}{*}{\rotatebox[origin=c]{90}{Ablation Study}}} 
& \multicolumn{2}{l|}{$\mathcal{G}_{\text{imb}}$}
& 0.477{\scriptsize(0.027)} & 0.375{\scriptsize(0.014)} 
& 0.432{\scriptsize(0.042)} & 0.637{\scriptsize(0.042)} &
& 0.273{\scriptsize(0.024)} & 0.215{\scriptsize(0.023)} 
& 0.248{\scriptsize(0.043)} & 0.401{\scriptsize(0.039)} \\
% \cmidrule{2-7}  \cmidrule{9-12} 
& \multicolumn{2}{l|}{ $\mathcal{G}_{\text{imb}}\cup\mathcal{G}_{\text{conf}}$ }
& 0.468{\scriptsize(0.007)} & 0.379{\scriptsize(0.012)} 
& 0.425{\scriptsize(0.013)} & 0.612{\scriptsize(0.028)} &
& \textbf{0.263}{\scriptsize(0.009)} & 0.219{\scriptsize(0.007)} 
& \textbf{0.236}{\scriptsize(0.017)} & 0.366{\scriptsize(0.020)} \\
& \multicolumn{2}{l|}{ $\mathcal{G}_{\text{imb}}\cup\mathcal{G}_{\text{conf}}$ (w/o  $\sigma$)}
& 0.480{\scriptsize(0.017)} & 0.380{\scriptsize(0.035)} 
& 0.440{\scriptsize(0.017)} & 0.630{\scriptsize(0.020)} &
& 0.269{\scriptsize(0.016)} & 0.219{\scriptsize(0.028)} 
& 0.249{\scriptsize(0.024)} & 0.368{\scriptsize(0.017)} \\
& \multicolumn{2}{l|}{ $\mathcal{G}_{\text{imb}}\cup\mathcal{G}_{\text{conf}}$ (w/o  $p$)}
& 0.475{\scriptsize(0.014)} & 0.369{\scriptsize(0.014)} 
& 0.446{\scriptsize(0.017)} & 0.618{\scriptsize(0.039)} &
& 0.267{\scriptsize(0.012)} & 0.210{\scriptsize(0.013)} 
& 0.251{\scriptsize(0.017)} & 0.372{\scriptsize(0.050)} \\
& \multicolumn{2}{l|}{  $\mathcal{G}_{\text{imb}}\cup\mathcal{H}_{\text{aug}}$ }
& 0.474{\scriptsize(0.010)} & \textbf{0.353}{\scriptsize(0.018)} 
& 0.450{\scriptsize(0.009)} & 0.623{\scriptsize(0.027)} &
& 0.272{\scriptsize(0.004)} & \textbf{0.202}{\scriptsize(0.012)} 
& 0.257{\scriptsize(0.011)} & 0.397{\scriptsize(0.034)} \\
& \multicolumn{2}{l|}{  $\mathcal{G}_{\text{imb}}\cup\mathcal{G}_{\text{conf}}\cup \mathcal{H}_{\text{aug}}$ }
& \textbf{0.457}{\scriptsize(0.015)} & 0.370{\scriptsize(0.022)}
&\textbf{ 0.411}{\scriptsize(0.011)} & \textbf{0.604}{\scriptsize(0.024)} &
& \textbf{0.263}{\scriptsize(0.016)} & 0.226{\scriptsize(0.021)} 
& 0.240{\scriptsize(0.015)} & \textbf{0.347}{\scriptsize(0.030)} \\
%%%%%%%%%%% D_a
\cmidrule{1-7} \cmidrule{9-12} 
\parbox[t]{5mm}{\multirow{9}{*}{\rotatebox[origin=c]{90}{$\textbf{z}_i$ and $\textbf{h}_j$ options in Mixup}}} 
&\multirow{3}{*}{\shortstack{\\  $\mathcal{G}_{\text{imb}}$} }
& $\mathcal{G}_{\text{imb}}$
& 0.466{\scriptsize(0.009)} & 0.374{\scriptsize(0.023)} 
& 0.430{\scriptsize(0.010)} & \textbf{0.604}{\scriptsize(0.032)}  &
& 0.266{\scriptsize(0.010)} & 0.214{\scriptsize(0.027)} 
& 0.242{\scriptsize(0.018)} & 0.379{\scriptsize(0.016)} \\
&& $\mathcal{G}_{\text{imb}} \cup \mathcal{G}_{\text{conf}} $
& 0.460{\scriptsize(0.016)} & 0.368{\scriptsize(0.026)} 
& 0.420{\scriptsize(0.018)} & 0.605{\scriptsize(0.026)} &
& 0.268{\scriptsize(0.017)} & 0.215{\scriptsize(0.023)} 
& 0.252{\scriptsize(0.022)} & 0.362{\scriptsize(0.016)} \\
&& $\mathcal{G}_{\text{imb}} \cup \mathcal{G}_{\text{unlbl}} $
& \textbf{0.457}{\scriptsize(0.015)} & 0.370{\scriptsize(0.022)} 
& \textbf{0.411}{\scriptsize(0.011)} & \textbf{0.604}{\scriptsize(0.024)}  &
& \textbf{0.263}{\scriptsize(0.016)} & 0.226{\scriptsize(0.021)} 
& \textbf{0.240}{\scriptsize(0.015)} & \textbf{0.347}{\scriptsize(0.030)}  \\
\cmidrule{2-7} \cmidrule{9-12} 
&\multirow{3}{*}{\shortstack{\\  $\mathcal{G}_{\text{imb}} \cup \mathcal{G}_{\text{conf}} $}  }
& $\mathcal{G}_{\text{imb}}$
& 0.469{\scriptsize(0.017)} & 0.369{\scriptsize(0.025)} 
& 0.432{\scriptsize(0.020)} & 0.615{\scriptsize(0.037)} &
& 0.260{\scriptsize(0.014)} & 0.204{\scriptsize(0.028)} 
& 0.248{\scriptsize(0.013)} & 0.358{\scriptsize(0.048)}  \\
&& $\mathcal{G}_{\text{imb}} \cup \mathcal{G}_{\text{conf}} $
& 0.466{\scriptsize(0.003)} & 0.376{\scriptsize(0.014)} 
& 0.425{\scriptsize(0.011)} & 0.610{\scriptsize(0.013)} &
& 0.261{\scriptsize(0.004)} & 0.204{\scriptsize(0.005)} 
& 0.242{\scriptsize(0.013)} & 0.370{\scriptsize(0.013)} \\
&& $\mathcal{G}_{\text{imb}} \cup \mathcal{G}_{\text{unlbl}} $
& 0.461{\scriptsize(0.010)} & \textbf{0.366}{\scriptsize(0.025)} 
& 0.424{\scriptsize(0.020)} & \textbf{0.604}{\scriptsize(0.026)} &
& 0.264{\scriptsize(0.015)} & 0.219{\scriptsize(0.027)} 
& 0.244{\scriptsize(0.017)} & 0.354{\scriptsize(0.036)} \\
\cmidrule{2-7}  \cmidrule{9-12} 
&\multirow{3}{*}{\shortstack{\\  $\mathcal{G}_{\text{imb}} \cup \mathcal{G}_{\text{unlbl}} $}  }
& $\mathcal{G}_{\text{imb}}$
& 0.472{\scriptsize(0.009)} & 0.369{\scriptsize(0.022)} 
& 0.435{\scriptsize(0.012)} & 0.623{\scriptsize(0.025)} &
& 0.266{\scriptsize(0.005)} & \textbf{0.202}{\scriptsize(0.015)} 
& 0.257{\scriptsize(0.012)} & 0.366{\scriptsize(0.016)} \\
&& $\mathcal{G}_{\text{imb}} \cup \mathcal{G}_{\text{conf}} $
& 0.476{\scriptsize(0.013)} & 0.387{\scriptsize(0.027)} 
& 0.426{\scriptsize(0.013)} & 0.630{\scriptsize(0.042)} &
& 0.271{\scriptsize(0.017)} & 0.211{\scriptsize(0.018)} 
& 0.253{\scriptsize(0.022)} & 0.382{\scriptsize(0.040)} \\
&& $\mathcal{G}_{\text{imb}} \cup \mathcal{G}_{\text{unlbl}} $
& 0.479{\scriptsize(0.026)} & 0.368{\scriptsize(0.012)} 
& 0.448{\scriptsize(0.033)} & 0.629{\scriptsize(0.047)} &
& 0.269{\scriptsize(0.016)} & 0.210{\scriptsize(0.010)} 
& 0.253{\scriptsize(0.023)} & 0.373{\scriptsize(0.033)} \\

\midrule
\multicolumn{12}{c}{\freesolv}  \\
\midrule
\parbox[t]{5mm}{\multirow{5}{*}{\rotatebox[origin=c]{90}{Ablation Study}}} 
& \multicolumn{2}{l|}{$\mathcal{G}_{\text{imb}}$}
& 0.619{\scriptsize(0.019)} & \textbf{0.525}{\scriptsize(0.022)} 
& 0.590{\scriptsize(0.035)} & 1.000{\scriptsize(0.072)} &
& 0.325{\scriptsize(0.040)} & 0.289{\scriptsize(0.006)} & 0.316{\scriptsize(0.062)} & 0.521{\scriptsize(0.084)} \\
& \multicolumn{2}{l|}{ $\mathcal{G}_{\text{imb}}\cup\mathcal{G}_{\text{conf}}$ }
& 0.568{\scriptsize(0.029)} & 0.538{\scriptsize(0.020)} 
& \textbf{0.520}{\scriptsize(0.045)} & 0.831{\scriptsize(0.132)} &
& 0.288{\scriptsize(0.031)} & 0.295{\scriptsize(0.037)} 
& 0.270{\scriptsize(0.037)} & 0.365{\scriptsize(0.088)} \\
& \multicolumn{2}{l|}{ $\mathcal{G}_{\text{imb}}\cup\mathcal{G}_{\text{conf}}$ (w/o  $\sigma$)}
& 0.660{\scriptsize(0.028)} & 0.574{\scriptsize(0.015)} 
& 0.650{\scriptsize(0.036)} & 0.941{\scriptsize(0.066)} &
& 0.325{\scriptsize(0.016)} & 0.302{\scriptsize(0.007)} 
& 0.319{\scriptsize(0.029)} & 0.437{\scriptsize(0.056)} \\
& \multicolumn{2}{l|}{ $\mathcal{G}_{\text{imb}}\cup\mathcal{G}_{\text{conf}}$ (w/o  $p$)}
& 0.604{\scriptsize(0.020)} & 0.557{\scriptsize(0.037)} 
& 0.560{\scriptsize(0.029)} & 0.903{\scriptsize(0.055)} &
& 0.293{\scriptsize(0.024)} & 0.307{\scriptsize(0.050)} 
& 0.260{\scriptsize(0.018)} & 0.416{\scriptsize(0.080)} \\
& \multicolumn{2}{l|}{  $\mathcal{G}_{\text{imb}}\cup\mathcal{H}_{\text{aug}}$ }
& 0.593{\scriptsize(0.045)} & 0.536{\scriptsize(0.033)} 
& 0.542{\scriptsize(0.067)} & 0.947{\scriptsize(0.062)} &
& 0.269{\scriptsize(0.022)} & 0.259{\scriptsize(0.037)} 
& 0.253{\scriptsize(0.050)} & 0.409{\scriptsize(0.033)} \\
& \multicolumn{2}{l|}{  $\mathcal{G}_{\text{imb}}\cup\mathcal{G}_{\text{conf}}\cup \mathcal{H}_{\text{aug}}$ }
& \textbf{0.563}{\scriptsize(0.026)} & 0.535{\scriptsize(0.038)}
& 0.528{\scriptsize(0.046)} & \textbf{0.777}{\scriptsize(0.061)} &
& \textbf{0.264}{\scriptsize(0.029)} & \textbf{0.286}{\scriptsize(0.013)}
& \textbf{0.244}{\scriptsize(0.046)} & \textbf{0.304}{\scriptsize(0.078)} \\
%%%%%%%%%%% D_a
\cmidrule{1-7} \cmidrule{9-12} 
\parbox[t]{5mm}{\multirow{9}{*}{\rotatebox[origin=c]{90}{$\textbf{z}_i$ and $\textbf{h}_j$ options in Mixup}}} 
&\multirow{3}{*}{\shortstack{\\ $\mathcal{G}_{\text{imb}}$} }
& $\mathcal{G}_{\text{imb}}$
& 0.572{\scriptsize(0.006)} & 0.528{\scriptsize(0.030)} 
& 0.531{\scriptsize(0.017)} & 0.852{\scriptsize(0.090)} &
& 0.289{\scriptsize(0.013)} & 0.299{\scriptsize(0.026)} 
& 0.265{\scriptsize(0.019)} & 0.370{\scriptsize(0.079)} \\
&& $\mathcal{G}_{\text{imb}} \cup \mathcal{G}_{\text{conf}} $
& 0.575{\scriptsize(0.017)} & 0.551{\scriptsize(0.018)} 
& 0.516{\scriptsize(0.034)} & 0.863{\scriptsize(0.071)} &
& 0.282{\scriptsize(0.014)} & 0.298{\scriptsize(0.015)} 
& 0.249{\scriptsize(0.012)} & 0.389{\scriptsize(0.058)}\\
&& $\mathcal{G}_{\text{imb}} \cup \mathcal{G}_{\text{unlbl}} $
& 0.563{\scriptsize(0.026)} & 0.535{\scriptsize(0.038)} 
& 0.528{\scriptsize(0.046)} & \textbf{0.777}{\scriptsize(0.061)} &
& 0.264{\scriptsize(0.029)} & 0.286{\scriptsize(0.013)} 
& 0.244{\scriptsize(0.046)} & \textbf{0.304}{\scriptsize(0.078)} \\
\cmidrule{2-7} \cmidrule{9-12} 
&\multirow{3}{*}{\shortstack{\\  $\mathcal{G}_{\text{imb}} \cup \mathcal{G}_{\text{conf}} $}  }
& $\mathcal{G}_{\text{imb}}$
& 0.568{\scriptsize(0.032)} & 0.535{\scriptsize(0.038)} 
& 0.513{\scriptsize(0.036)} & 0.867{\scriptsize(0.083)} &
& 0.267{\scriptsize(0.019)} & 0.285{\scriptsize(0.020)} 
& 0.235{\scriptsize(0.026)} & 0.357{\scriptsize(0.035)} \\
&& $\mathcal{G}_{\text{imb}} \cup \mathcal{G}_{\text{conf}} $
& 0.577{\scriptsize(0.021)} & 0.537{\scriptsize(0.052)} 
& 0.522{\scriptsize(0.012)} & 0.896{\scriptsize(0.020)} &
& 0.280{\scriptsize(0.018)} & 0.301{\scriptsize(0.040)} 
& 0.246{\scriptsize(0.018)} & 0.374{\scriptsize(0.048)} \\
&& $\mathcal{G}_{\text{imb}} \cup \mathcal{G}_{\text{unlbl}} $
& 0.565{\scriptsize(0.027)} & \textbf{0.518}{\scriptsize(0.034)} 
& 0.522{\scriptsize(0.034)} & 0.864{\scriptsize(0.110)} &
& \textbf{0.262}{\scriptsize(0.024)} & \textbf{0.255}{\scriptsize(0.026)} 
& 0.247{\scriptsize(0.022)} & 0.360{\scriptsize(0.086)} \\
\cmidrule{2-7}  \cmidrule{9-12} 
&\multirow{3}{*}{\shortstack{\\  $\mathcal{G}_{\text{imb}} \cup \mathcal{G}_{\text{unlbl}} $}  }
& $\mathcal{G}_{\text{imb}}$
& 0.621{\scriptsize(0.053)} & 0.555{\scriptsize(0.044)} 
& 0.587{\scriptsize(0.063)} & 0.939{\scriptsize(0.176)} &
& 0.327{\scriptsize(0.048)} & 0.321{\scriptsize(0.024)} 
& 0.304{\scriptsize(0.059)} & 0.473{\scriptsize(0.105)}  \\
&& $\mathcal{G}_{\text{imb}} \cup \mathcal{G}_{\text{conf}} $
& 0.598{\scriptsize(0.042)} & 0.552{\scriptsize(0.029)} 
& 0.545{\scriptsize(0.040)} & 0.924{\scriptsize(0.097)} &
& 0.311{\scriptsize(0.040)} & 0.300{\scriptsize(0.051)} 
& 0.295{\scriptsize(0.040)} & 0.428{\scriptsize(0.067)}\\
&& $\mathcal{G}_{\text{imb}} \cup \mathcal{G}_{\text{unlbl}} $
& \textbf{0.559}{\scriptsize(0.023)} & \textbf{0.518}{\scriptsize(0.023)} 
& \textbf{0.503}{\scriptsize(0.016)} & 0.882{\scriptsize(0.081)} &
& 0.266{\scriptsize(0.017)} & 0.278{\scriptsize(0.029)} 
&\textbf{0.229}{\scriptsize(0.010)} & 0.410{\scriptsize(0.047)} \\

\bottomrule[1.5pt]
\end{tabular}}
\end{table*}
\begin{table*}[ht]
\centering
\caption{Complete results of ablation study and mixup options ({MAE $\downarrow$} and GM $\downarrow$) on three polymer datasets. The best mean is \textbf{bolded}. For the label-anchored mixup options, the first column is the source of $\textbf{z}_i$ and the second column is the source of $\textbf{h}_j$.}
\label{table:ablation_all_plym}
\setlength{\tabcolsep}{2pt}
\resizebox{0.96\linewidth}{!}{\begin{tabular}{@{}c|ll|ccccccccc@{}}\toprule[1.2pt]
\multicolumn{3}{c}{} & \multicolumn{4}{c}{{MAE $\downarrow$}} & \phantom{abc}& \multicolumn{4}{c}{{GM $\downarrow$}} \\
\cmidrule{4-7} \cmidrule{9-12} 
\multicolumn{3}{c}{} & All & Many-shot & Med.-shot & Few-shot &&  All & Many-shot & Med.-shot & Few-shot \\
\midrule
\multicolumn{12}{c}{\meltTemp}  \\
\midrule
\parbox[t]{5mm}{\multirow{5}{*}{\rotatebox[origin=c]{90}{Ablation Study}}} 
& \multicolumn{2}{l|}{$\mathcal{G}_{\text{imb}}$}
& 41.1{\scriptsize(1.4)} & 32.7{\scriptsize(2.7)} 
& 30.3{\scriptsize(0.9)} & 57.4{\scriptsize(2.1)} &
& 21.9{\scriptsize(0.5)} & 19.0{\scriptsize(1.9)} 
& \textbf{14.4}{\scriptsize(0.9)} & 34.9{\scriptsize(1.8)} \\
& \multicolumn{2}{l|}{ $\mathcal{G}_{\text{imb}}\cup\mathcal{G}_{\text{conf}}$ }
& 40.0{\scriptsize(0.7)} & 32.7{\scriptsize(1.9)} 
& 31.4{\scriptsize(1.3)} & 53.6{\scriptsize(1.9)} &
& 21.3{\scriptsize(1.0)} & \textbf{17.7}{\scriptsize(1.6)} 
& 15.8{\scriptsize(1.3)} & 32.4{\scriptsize(2.6)} \\
& \multicolumn{2}{l|}{ $\mathcal{G}_{\text{imb}}\cup\mathcal{G}_{\text{conf}}$ (w/o  $\sigma$)}
& 41.2{\scriptsize(1.1)} & 33.0{\scriptsize(1.6)} 
& 32.1{\scriptsize(0.7)} & 56.2{\scriptsize(1.7)} &
& 22.2{\scriptsize(1.1)} & 18.9{\scriptsize(1.4)} 
& 15.9{\scriptsize(1.3)} & 33.7{\scriptsize(1.2)} \\
& \multicolumn{2}{l|}{ $\mathcal{G}_{\text{imb}}\cup\mathcal{G}_{\text{conf}}$ (w/o  $p$)}
& 40.3{\scriptsize(1.0)} & 32.5{\scriptsize(1.4)} 
& 31.3{\scriptsize(1.1)} & 54.7{\scriptsize(1.8)} &
& 21.7{\scriptsize(1.1)} & 18.2{\scriptsize(1.0)} 
& 15.2{\scriptsize(1.0)} & 33.9{\scriptsize(2.2)} \\
& \multicolumn{2}{l|}{  $\mathcal{G}_{\text{imb}}\cup\mathcal{H}_{\text{aug}}$ }
& 40.4{\scriptsize(0.4)} & 32.5{\scriptsize(1.5)} 
& \textbf{30.2}{\scriptsize(1.3)} & 55.9{\scriptsize(1.1)} &
& 21.9{\scriptsize(0.8)} & 19.7{\scriptsize(1.8)} 
& \textbf{14.4}{\scriptsize(0.8)} & 34.0{\scriptsize(0.9)} \\
& \multicolumn{2}{l|}{  $\mathcal{G}_{\text{imb}}\cup\mathcal{G}_{\text{conf}}\cup \mathcal{H}_{\text{aug}}$ }
& \textbf{38.9}{\scriptsize(0.7)} & \textbf{31.7}{\scriptsize(0.3)} 
& 31.5{\scriptsize(1.1)} & \textbf{51.4}{\scriptsize(1.6)} &
& \textbf{21.1}{\scriptsize(1.2)} & 18.5{\scriptsize(0.5)} 
& 15.9{\scriptsize(1.4)} & \textbf{30.2}{\scriptsize(1.9)} \\
%%%%%%%%%%% D_a
\cmidrule{1-7} \cmidrule{9-12} 
\parbox[t]{5mm}{\multirow{10}{*}{\rotatebox[origin=c]{90}{$\textbf{z}_i$ and $\textbf{h}_j$ options in Mixup}}} 
&\multirow{3}{*}{\shortstack{ \\ $\mathcal{G}_{\text{imb}}$} }
& $\mathcal{G}_{\text{imb}}$
& 39.9{\scriptsize(1.0)} & 32.7{\scriptsize(1.2)} 
& 30.9{\scriptsize(1.4)} & 53.8{\scriptsize(1.8)} &
& 21.4{\scriptsize(0.6)} & 18.8{\scriptsize(0.6)} 
& \textbf{14.6}{\scriptsize(0.9)} & 32.8{\scriptsize(1.2)} \\
&& $\mathcal{G}_{\text{imb}} \cup \mathcal{G}_{\text{conf}} $
& 40.3{\scriptsize(2.0)} & 32.5{\scriptsize(1.5)} 
& \textbf{30.8}{\scriptsize(0.9)} & 55.2{\scriptsize(4.9)} &
& 21.9{\scriptsize(1.9)} & 19.2{\scriptsize(1.7)} 
& 15.0{\scriptsize(1.0)} & 33.6{\scriptsize(5.3)} \\
&& $\mathcal{G}_{\text{imb}} \cup \mathcal{G}_{\text{unlbl}} $
& \textbf{38.9}{\scriptsize(0.7)} & \textbf{31.7}{\scriptsize(0.3)} & 31.5{\scriptsize(1.1)} & \textbf{51.4}{\scriptsize(1.6)} && 
\textbf{21.1}{\scriptsize(1.2)} & 18.5{\scriptsize(0.5)} & 15.9{\scriptsize(1.4)} & \textbf{30.2}{\scriptsize(1.9)} \\
\cmidrule{2-7} \cmidrule{9-12} 
&\multirow{3}{*}{\shortstack{ \\  $\mathcal{G}_{\text{imb}} \cup \mathcal{G}_{\text{conf}} $}  }
& $\mathcal{G}_{\text{imb}}$
& 40.5{\scriptsize(1.0)} & 32.0{\scriptsize(1.2)} 
& \textbf{30.8}{\scriptsize(1.2)} & 56.1{\scriptsize(1.7)} &
& 21.7{\scriptsize(1.3)} & 18.4{\scriptsize(1.0)} 
& 14.8{\scriptsize(1.6)} & 34.5{\scriptsize(1.9)} \\
&& $\mathcal{G}_{\text{imb}} \cup \mathcal{G}_{\text{conf}} $
& 40.5{\scriptsize(1.2)} & 33.2{\scriptsize(1.8)} 
& 31.7{\scriptsize(0.5)} & 54.3{\scriptsize(1.7)} &
& 21.4{\scriptsize(1.0)} & \textbf{18.3}{\scriptsize(1.7)} 
& 15.1{\scriptsize(0.9)} & 32.7{\scriptsize(1.2)}  \\
&& $\mathcal{G}_{\text{imb}} \cup \mathcal{G}_{\text{unlbl}} $
& 40.1{\scriptsize(0.6)} & 32.3{\scriptsize(1.8)} 
& 31.5{\scriptsize(0.9)} & 54.3{\scriptsize(1.2)} &
& 21.8{\scriptsize(0.8)} & 18.4{\scriptsize(1.4)} 
& 15.9{\scriptsize(1.4)} & 33.1{\scriptsize(1.4)}\\
\cmidrule{2-7}  \cmidrule{9-12} 
&\multirow{3}{*}{\shortstack{ \\  $\mathcal{G}_{\text{imb}} \cup \mathcal{G}_{\text{unlbl}} $}  }
& $\mathcal{G}_{\text{imb}}$
& 40.7{\scriptsize(1.4)} & \textbf{31.7}{\scriptsize(1.1)} 
& 31.7{\scriptsize(1.6)} & 56.3{\scriptsize(4.5)} &
& 21.9{\scriptsize(0.9)} & \textbf{18.3}{\scriptsize(0.5)} 
& 15.0{\scriptsize(1.4)} & 35.4{\scriptsize(3.9)} \\
&& $\mathcal{G}_{\text{imb}} \cup \mathcal{G}_{\text{conf}} $
& 40.5{\scriptsize(1.7)} & 32.3{\scriptsize(2.8)} 
& 31.2{\scriptsize(1.5)} & 55.4{\scriptsize(3.5)} &
& 22.0{\scriptsize(1.0)} & 18.7{\scriptsize(2.1)} 
& 15.4{\scriptsize(1.7)} & 34.4{\scriptsize(3.4)} \\
&& $\mathcal{G}_{\text{imb}} \cup \mathcal{G}_{\text{unlbl}} $
& 40.9{\scriptsize(1.4)} & 33.3{\scriptsize(1.8)} 
& 31.6{\scriptsize(1.6)} & 55.4{\scriptsize(2.9)} &
& 22.2{\scriptsize(1.1)} & 19.4{\scriptsize(1.4)} 
& 15.3{\scriptsize(1.8)} & 34.0{\scriptsize(0.6)} \\

\midrule
\multicolumn{12}{c}{\density (scaled:$ \times 1,000$)}  \\
\midrule
\parbox[t]{5mm}{\multirow{5}{*}{\rotatebox[origin=c]{90}{Ablation Study}}} 
& \multicolumn{2}{l|}{$\mathcal{G}_{\text{imb}}$}
& 56.8{\scriptsize(2.1)} & 49.4{\scriptsize(4.8)} 
& 46.7{\scriptsize(2.3)} & 72.1{\scriptsize(2.1)} &
& 29.9{\scriptsize(2.1)} & 27.4{\scriptsize(2.3)} 
& 25.6{\scriptsize(3.6)} & 37.2{\scriptsize(1.3)} \\
& \multicolumn{2}{l|}{ $\mathcal{G}_{\text{imb}}\cup\mathcal{G}_{\text{conf}}$}
& 54.5{\scriptsize(0.6)} & 49.0{\scriptsize(2.6)} 
& 42.9{\scriptsize(2.1)} & 69.3{\scriptsize(0.8)}&
& 27.3{\scriptsize(0.8)} & 26.3{\scriptsize(0.9)} 
& \textbf{21.5}{\scriptsize(1.4)} & 34.8{\scriptsize(2.6)} \\
& \multicolumn{2}{l|}{ $\mathcal{G}_{\text{imb}}\cup\mathcal{G}_{\text{conf}}$ (w/o  $\sigma$)}
& 58.0{\scriptsize(1.4)} & 47.5{\scriptsize(2.2)} 
& 45.7{\scriptsize(3.2)} & 77.7{\scriptsize(2.0)} &
& 29.0{\scriptsize(1.4)} & 27.1{\scriptsize(2.8)} 
& 23.1{\scriptsize(2.3)} & 38.0{\scriptsize(3.1)} \\
& \multicolumn{2}{l|}{ $\mathcal{G}_{\text{imb}}\cup\mathcal{G}_{\text{conf}}$ (w/o  $p$)}
& 55.9{\scriptsize(4.8)} & 50.4{\scriptsize(10.0)} 
& 44.3{\scriptsize(3.1)} & 70.8{\scriptsize(4.0)} &
& 29.1{\scriptsize(3.6)} & 29.4{\scriptsize(9.0)} 
& 23.2{\scriptsize(2.6)} & 35.9{\scriptsize(3.9)} \\
& \multicolumn{2}{l|}{  $\mathcal{G}_{\text{imb}}\cup\mathcal{H}_{\text{aug}}$ }
& 55.4{\scriptsize(3.2)} & 50.5{\scriptsize(5.6)}
& 44.3{\scriptsize(1.0)} & 69.2{\scriptsize(4.1)} &
& 29.1{\scriptsize(3.8)} & 28.0{\scriptsize(4.8)} 
& 25.0{\scriptsize(3.0)} & 34.7{\scriptsize(4.8)} \\
& \multicolumn{2}{l|}{  $\mathcal{G}_{\text{imb}}\cup\mathcal{G}_{\text{conf}}\cup \mathcal{H}_{\text{aug}}$ }
& \textbf{53.0}{\scriptsize(0.5)} & \textbf{45.4}{\scriptsize(1.7)} 
& \textbf{42.5}{\scriptsize(2.8)} & \textbf{68.6}{\scriptsize(2.6)} &
& \textbf{26.6}{\scriptsize(0.4)} & \textbf{24.0}{\scriptsize(2.2)} 
& 23.0{\scriptsize(1.3)} & \textbf{33.4}{\scriptsize(3.0)} \\
%%%%%%%%%%% D_a
\cmidrule{1-7} \cmidrule{9-12} 
\parbox[t]{5mm}{\multirow{10}{*}{\rotatebox[origin=c]{90}{$\textbf{z}_i$ and $\textbf{h}_j$ options in Mixup}}} 
&\multirow{3}{*}{\shortstack{ \\  $\mathcal{G}_{\text{imb}}$} }
& $\mathcal{G}_{\text{imb}}$
& 55.6{\scriptsize(2.6)} & 47.1{\scriptsize(4.0)} 
& 44.1{\scriptsize(3.0)} & 73.0{\scriptsize(3.1)} & 
& 29.1{\scriptsize(1.5)} & 25.9{\scriptsize(1.8)} 
& 24.4{\scriptsize(2.4)} & 37.7{\scriptsize(1.7)} \\
&& $\mathcal{G}_{\text{imb}} \cup \mathcal{G}_{\text{conf}} $
& 54.2{\scriptsize(0.4)} & 46.2{\scriptsize(2.9)} 
& 42.9{\scriptsize(2.7)} & 71.0{\scriptsize(1.0)} & 
& 27.4{\scriptsize(1.1)} & 25.1{\scriptsize(2.6)} 
& 22.3{\scriptsize(1.2)} & 35.6{\scriptsize(2.4)} \\
&& $\mathcal{G}_{\text{imb}} \cup \mathcal{G}_{\text{unlbl}} $
& \textbf{53.0}{\scriptsize(0.5)} & \textbf{45.4}{\scriptsize(1.7)} & 42.5{\scriptsize(2.8)} & \textbf{68.6}{\scriptsize(2.6)} &&
\textbf{26.6}{\scriptsize(0.4)} & \textbf{24.0}{\scriptsize(2.2)} & 23.0{\scriptsize(1.3)} & 33.4{\scriptsize(3.0)} \\
\cmidrule{2-7} \cmidrule{9-12} 
&\multirow{3}{*}{\shortstack{ \\  $\mathcal{G}_{\text{imb}} \cup \mathcal{G}_{\text{conf}} $}  }
& $\mathcal{G}_{\text{imb}}$
& 58.7{\scriptsize(3.5)} & 52.2{\scriptsize(3.8)} 
& 45.4{\scriptsize(1.0)} & 75.9{\scriptsize(6.6)} &
& 32.4{\scriptsize(2.7)} & 30.9{\scriptsize(3.6)} 
& 25.1{\scriptsize(1.4)} & 42.5{\scriptsize(5.4)} \\
&& $\mathcal{G}_{\text{imb}} \cup \mathcal{G}_{\text{conf}} $
& 56.3{\scriptsize(1.9)} & 49.1{\scriptsize(4.7)} 
& 43.4{\scriptsize(2.2)} & 73.8{\scriptsize(5.3)} & 
& 28.8{\scriptsize(2.5)} & 27.2{\scriptsize(3.5)} 
& 22.5{\scriptsize(1.6)} & 37.9{\scriptsize(4.8)} \\
&& $\mathcal{G}_{\text{imb}} \cup \mathcal{G}_{\text{unlbl}} $
& 54.7{\scriptsize(0.9)} & 50.3{\scriptsize(0.7)} 
& 42.0{\scriptsize(0.5)} & 69.5{\scriptsize(2.7)} & 
& 27.8{\scriptsize(0.8)} & 29.4{\scriptsize(1.6)} 
& 21.6{\scriptsize(2.0)} & \textbf{33.2}{\scriptsize(1.1)} \\
\cmidrule{2-7}  \cmidrule{9-12} 
&\multirow{3}{*}{\shortstack{ \\  $\mathcal{G}_{\text{imb}} \cup \mathcal{G}_{\text{unlbl}} $}  }
& $\mathcal{G}_{\text{imb}}$
& 58.8{\scriptsize(9.2)} & 53.9{\scriptsize(10.8)} 
& 45.9{\scriptsize(7.2)} & 74.3{\scriptsize(9.9)} & 
& 30.9{\scriptsize(7.8)} & 30.1{\scriptsize(7.6)} 
& 25.7{\scriptsize(6.9)} & 37.2{\scriptsize(9.0)} \\
&& $\mathcal{G}_{\text{imb}} \cup \mathcal{G}_{\text{conf}} $
& 55.9{\scriptsize(3.1)} & 49.9{\scriptsize(4.2)} 
& 43.2{\scriptsize(4.1)} & 72.2{\scriptsize(4.4)} & 
& 27.9{\scriptsize(3.1)} & 26.6{\scriptsize(3.2)} 
& 22.5{\scriptsize(2.7)} & 35.4{\scriptsize(7.2)} \\
&& $\mathcal{G}_{\text{imb}} \cup \mathcal{G}_{\text{unlbl}} $
& 55.0{\scriptsize(1.8)} & 49.0{\scriptsize(5.1)} 
& \textbf{41.2}{\scriptsize(0.6)} & 72.3{\scriptsize(3.1)} & 
& 27.5{\scriptsize(1.8)} & 27.3{\scriptsize(1.9)} 
& \textbf{21.0}{\scriptsize(1.3)} & 35.1{\scriptsize(3.9)} \\

\midrule
\multicolumn{12}{c}{\oxygen}  \\
\midrule
\parbox[t]{5mm}{\multirow{5}{*}{\rotatebox[origin=c]{90}{Ablation Study}}} 
& \multicolumn{2}{l|}{$\mathcal{G}_{\text{imb}}$}
& 160.0{\scriptsize(24.7)} & 10.2{\scriptsize(10.1)} 
& 11.4{\scriptsize(1.2)} & 400.8{\scriptsize(57.9)}  &
& 6.4{\scriptsize(0.6)} & 2.3{\scriptsize(0.4)} 
& 4.1{\scriptsize(0.6)} & 24.8{\scriptsize(4.8)} \\
& \multicolumn{2}{l|}{ $\mathcal{G}_{\text{imb}}\cup\mathcal{G}_{\text{conf}}$ }
& 158.2{\scriptsize(8.8)} & 5.4{\scriptsize(2.8)} & 13.8{\scriptsize(2.1)} & 399.2{\scriptsize(22.3)} &
& 6.0{\scriptsize(0.5)} & 2.1{\scriptsize(0.5)} & 3.6{\scriptsize(0.9)} & 24.5{\scriptsize(3.5)} \\
& \multicolumn{2}{l|}{ $\mathcal{G}_{\text{imb}}\cup\mathcal{G}_{\text{conf}}$ (w/o  $\sigma$)}
& 180.2{\scriptsize(4.0)} & 4.5{\scriptsize(2.0)} 
& 15.1{\scriptsize(4.8)} & 456.9{\scriptsize(11.9)} &
& 7.8{\scriptsize(1.2)} & 2.1{\scriptsize(0.6)} 
& 5.3{\scriptsize(0.6)} & 37.0{\scriptsize(3.6)} \\
& \multicolumn{2}{l|}{ $\mathcal{G}_{\text{imb}}\cup\mathcal{G}_{\text{conf}}$ (w/o  $p$)}
& 168.4{\scriptsize(22.4)} & 7.0{\scriptsize(6.4)} 
& 14.8{\scriptsize(4.3)} & 423.7{\scriptsize(52.7)} &
& 7.0{\scriptsize(1.4)} & 2.1{\scriptsize(0.5)} 
& 4.4{\scriptsize(1.6)} & 31.7{\scriptsize(5.9)} \\
& \multicolumn{2}{l|}{  $\mathcal{G}_{\text{imb}}\cup\mathcal{H}_{\text{aug}}$ }
& 157.7{\scriptsize(21.7)} & \textbf{3.8}{\scriptsize(0.7)} 
& 13.3{\scriptsize(1.8)} & 399.9{\scriptsize(57.3)}  &
& 5.9{\scriptsize(0.4)} & \textbf{1.9}{\scriptsize(0.2)} 
& 3.6{\scriptsize(0.9)} & 25.3{\scriptsize(1.7)} \\
& \multicolumn{2}{l|}{  $\mathcal{G}_{\text{imb}}\cup\mathcal{G}_{\text{conf}}\cup \mathcal{H}_{\text{aug}}$ }
& \textbf{150.9{}\scriptsize(17.8)} & \textbf{3.8}{\scriptsize(1.1)} 
& \textbf{12.2}{\scriptsize(0.6)} & \textbf{382.8}{\scriptsize(46.9)} &
& \textbf{5.8}{\scriptsize(0.4)} & 2.1{\scriptsize(0.7)} 
& \textbf{3.3}{\scriptsize(0.8)} & \textbf{24.4}{\scriptsize(6.8)} \\
%%%%%%%%%%% D_a
\cmidrule{1-7} \cmidrule{9-12} 
\parbox[t]{5mm}{\multirow{10}{*}{\rotatebox[origin=c]{90}{$\textbf{z}_i$ and $\textbf{h}_j$ options in Mixup}}} 
& \multirow{3}{*}{\shortstack{ \\ $\mathcal{G}_{\text{imb}}$} }
& $\mathcal{G}_{\text{imb}}$
& 165.5{\scriptsize(12.2)} & 4.7{\scriptsize(1.7)} & 16.5{\scriptsize(7.2)} & 417.4{\scriptsize(31.1)} &
& 6.0{\scriptsize(0.7)} & \textbf{1.9}{\scriptsize(0.5)} 
& 3.6{\scriptsize(0.3)} & 25.8{\scriptsize(3.2)} \\
&& $\mathcal{G}_{\text{imb}} \cup \mathcal{G}_{\text{conf}} $
& 158.1{\scriptsize(17.0)} & 4.1{\scriptsize(0.7)} & \textbf{11.3}{\scriptsize(0.7)} & 401.9{\scriptsize(45.1)} &
& 6.8{\scriptsize(1.3)} & 2.3{\scriptsize(0.3)} 
& 4.2{\scriptsize(0.9)} & 27.7{\scriptsize(9.5)} \\
&& $\mathcal{G}_{\text{imb}} \cup \mathcal{G}_{\text{unlbl}} $
& \textbf{150.9}{\scriptsize(17.8)} & 3.8{\scriptsize(1.1)} 
& 12.2{\scriptsize(0.6)} & \textbf{382.8}{\scriptsize(46.9)} &
& \textbf{5.8}{\scriptsize(0.4)} & 2.1{\scriptsize(0.7)} & 
\textbf{3.3}{\scriptsize(0.8)} & \textbf{24.4}{\scriptsize(6.8)} \\
\cmidrule{2-7} \cmidrule{9-12} 
&\multirow{3}{*}{\shortstack{ \\  $\mathcal{G}_{\text{imb}} \cup \mathcal{G}_{\text{conf}} $}  }
& $\mathcal{G}_{\text{imb}}$
& 166.0{\scriptsize(18.2)} & 11.9{\scriptsize(11.3)} 
& 12.6{\scriptsize(0.9)} & 414.0{\scriptsize(52.6)}  &
& 6.5{\scriptsize(0.6)} & 2.1{\scriptsize(0.3)} 
& 3.7{\scriptsize(0.9)} & 28.8{\scriptsize(3.1)} \\
&& $\mathcal{G}_{\text{imb}} \cup \mathcal{G}_{\text{conf}} $
& 158.8{\scriptsize(8.4)} & 7.7{\scriptsize(8.9)} 
& 15.4{\scriptsize(7.8)} & 397.5{\scriptsize(15.4)}  &
& 6.8{\scriptsize(1.5)} & 2.0{\scriptsize(0.7)} 
& 4.4{\scriptsize(1.4)} & 30.2{\scriptsize(2.2)}  \\
&& $\mathcal{G}_{\text{imb}} \cup \mathcal{G}_{\text{unlbl}} $
& 169.5{\scriptsize(56.1)} & 4.5{\scriptsize(1.2)} & 12.7{\scriptsize(1.8)} & 430.4{\scriptsize(145.0)}  &
& 7.9{\scriptsize(2.1)} & 2.3{\scriptsize(0.4)} 
& 5.4{\scriptsize(2.0)} & 35.1{\scriptsize(11.3)}  \\
\cmidrule{2-7}  \cmidrule{9-12} 
&\multirow{3}{*}{\shortstack{ \\  $\mathcal{G}_{\text{imb}} \cup \mathcal{G}_{\text{unlbl}} $}  }
& $\mathcal{G}_{\text{imb}}$
& 173.1{\scriptsize(30.3)} & \textbf{3.7}{\scriptsize(0.4)} 
& 13.5{\scriptsize(1.4)} & 440.0{\scriptsize(79.3)} &
& 6.6{\scriptsize(1.3)} & \textbf{1.9}{\scriptsize(0.2)} 
& 4.0{\scriptsize(1.6)} & 31.1{\scriptsize(5.8)} \\
&& $\mathcal{G}_{\text{imb}} \cup \mathcal{G}_{\text{conf}} $
& 174.5{\scriptsize(9.3)} & 8.1{\scriptsize(3.3)} 
& 11.9{\scriptsize(0.9)} & 440.4{\scriptsize(25.5)} &
& 7.6{\scriptsize(2.2)} & 2.8{\scriptsize(0.8)} 
& 4.5{\scriptsize(2.0)} & 29.4{\scriptsize(7.0)}  \\
&& $\mathcal{G}_{\text{imb}} \cup \mathcal{G}_{\text{unlbl}} $
& 156.3{\scriptsize(20.5)} & 8.2{\scriptsize(2.9)} 
& 12.9{\scriptsize(0.9)} & 392.3{\scriptsize(50.6)}  &
& 9.8{\scriptsize(2.5)} & 3.8{\scriptsize(1.5)} 
& 6.1{\scriptsize(1.7)} & 34.8{\scriptsize(6.8)} \\
\bottomrule[1.5pt]
\end{tabular}}
\end{table*}
\begin{table*}[ht!]
\centering
\caption{Investigating the effect of regression confidence measurements ({MAE $\downarrow$} and GM $\downarrow$). The best mean is \textbf{bolded}.}
\label{table:uncertainty_all}
\setlength{\tabcolsep}{2pt}
\resizebox{0.98\linewidth}{!}{\begin{tabular}{@{}llccccccccc@{}}\toprule[1.2pt]
\multicolumn{2}{c}{} & \multicolumn{4}{c}{{MAE $\downarrow$}} & \phantom{abc}& \multicolumn{4}{c}{{GM $\downarrow$}} \\
\cmidrule{3-6} \cmidrule{8-11} 
\multicolumn{2}{c}{} & All & Many-shot & Med.-shot & Few-shot &&  All & Many-shot & Med.-shot & Few-shot \\
\midrule[1.2pt]
\multirow{5}{*}{\lipo}
& \textsc{Simple} &
0.481{\scriptsize(0.010)} & 0.389{\scriptsize(0.007)} & 0.440{\scriptsize(0.013)} & 0.603{\scriptsize(0.023)} &&
0.297{\scriptsize(0.014)} & 0.239{\scriptsize(0.006)} & 0.275{\scriptsize(0.019)} & 0.388{\scriptsize(0.026)} \\
& \textsc{Dropout} & 
0.450{\scriptsize(0.026)} & \textbf{0.365}{\scriptsize(0.031)} & \textbf{0.420}{\scriptsize(0.022)} & 0.555{\scriptsize(0.037)} && 
0.277{\scriptsize(0.017)} & 0.230{\scriptsize(0.020)} & 0.263{\scriptsize(0.011)} & 0.348{\scriptsize(0.044)} \\
& \textsc{Certi} &  
0.452{\scriptsize(0.011)} & 0.384{\scriptsize(0.018)} & 0.433{\scriptsize(0.013)} & \textbf{0.532}{\scriptsize(0.010)} && 
0.276{\scriptsize(0.009)} & 0.239{\scriptsize(0.017)} & 0.267{\scriptsize(0.015)} & \textbf{0.324}{\scriptsize(0.016)} \\
& \textsc{DER} & 
1.026{\scriptsize(0.033)} & 0.604{\scriptsize(0.035)} & 0.760{\scriptsize(0.016)} & 1.672{\scriptsize(0.111)} && 
0.688{\scriptsize(0.026)} & 0.417{\scriptsize(0.016)} & 0.528{\scriptsize(0.015)} & 1.405{\scriptsize(0.152)} \\
& \textsc{GRation} & 
\textbf{0.448}{\scriptsize(0.006)} & 0.371{\scriptsize(0.004)} & 0.421{\scriptsize(0.012)} & 0.543{\scriptsize(0.016)} && 
\textbf{0.270}{\scriptsize(0.002)} & \textbf{0.228}{\scriptsize(0.009)} & \textbf{0.255}{\scriptsize(0.008)} & 0.333{\scriptsize(0.015)} \\
\midrule

\multirow{5}{*}{\esol}
& \textsc{Simple} & 
0.499{\scriptsize(0.016)} & 0.397{\scriptsize(0.023)} & 0.457{\scriptsize(0.018)} & 0.656{\scriptsize(0.033)} &&
0.290{\scriptsize(0.017)} & 0.238{\scriptsize(0.023)} & 0.258{\scriptsize(0.020)} & 0.415{\scriptsize(0.025)} \\
& \textsc{Dropout} & 
0.483{\scriptsize(0.011)} & 0.381{\scriptsize(0.027)} & 0.443{\scriptsize(0.018)} & 0.636{\scriptsize(0.027)} &&
0.279{\scriptsize(0.017)} & 0.220{\scriptsize(0.019)} & 0.261{\scriptsize(0.026)} & 0.391{\scriptsize(0.032)} \\
& \textsc{Certi} &  
0.487{\scriptsize(0.030)} & 0.389{\scriptsize(0.039)} & \textbf{0.439}{\scriptsize(0.024)} & 0.647{\scriptsize(0.043)} && 
0.274{\scriptsize(0.018)} & 0.221{\scriptsize(0.033)} & \textbf{0.246}{\scriptsize(0.013)} & 0.396{\scriptsize(0.025)} \\
& \textsc{DER} & 
0.918{\scriptsize(0.135)} & 0.776{\scriptsize(0.086)} & 0.826{\scriptsize(0.098)} & 1.182{\scriptsize(0.245)} && 
0.619{\scriptsize(0.089)} & 0.525{\scriptsize(0.074)} & 0.567{\scriptsize(0.063)} & 0.829{\scriptsize(0.180)} \\
& \textsc{GRation} & 
\textbf{0.475}{\scriptsize(0.014)} & \textbf{0.369}{\scriptsize(0.014)} & 0.446{\scriptsize(0.017)} & \textbf{0.618}{\scriptsize(0.039)} && 
\textbf{0.267}{\scriptsize(0.012)} & \textbf{0.210}{\scriptsize(0.013)} & 0.251{\scriptsize(0.017)} & \textbf{0.372}{\scriptsize(0.050)} \\
\midrule

\multirow{5}{*}{\freesolv}
& \textsc{Simple} & 
0.697{\scriptsize(0.056)} & 0.616{\scriptsize(0.025)} & 0.663{\scriptsize(0.033)} & 1.054{\scriptsize(0.260)} && 
0.327{\scriptsize(0.036)} & 0.319{\scriptsize(0.028)} & 0.297{\scriptsize(0.017)} & 0.527{\scriptsize(0.206)} \\
& \textsc{Dropout} & 
0.639{\scriptsize(0.013)} & 0.578{\scriptsize(0.060)} & 0.589{\scriptsize(0.017)} & 1.005{\scriptsize(0.140)} && 
0.301{\scriptsize(0.018)} & \textbf{0.274}{\scriptsize(0.047)} & 0.299{\scriptsize(0.038)} & 0.433{\scriptsize(0.040)} \\
& \textsc{Certi} &  
0.654{\scriptsize(0.049)} & 0.589{\scriptsize(0.046)} & 0.611{\scriptsize(0.053)} & 0.999{\scriptsize(0.130)} && 
0.326{\scriptsize(0.038)} & 0.332{\scriptsize(0.040)} & 0.292{\scriptsize(0.044)} & 0.485{\scriptsize(0.095)} \\
& \textsc{DER} & 
1.483{\scriptsize(0.174)} & 1.180{\scriptsize(0.162)} & 1.450{\scriptsize(0.188)} & 2.480{\scriptsize(0.373)} && 
0.949{\scriptsize(0.131)} & 0.856{\scriptsize(0.159)} & 0.883{\scriptsize(0.183)} & 1.828{\scriptsize(0.386)} \\
& \textsc{GRation} & 
\textbf{0.604}{\scriptsize(0.020)} & \textbf{0.557}{\scriptsize(0.037)} & \textbf{0.560}{\scriptsize(0.029)} & \textbf{0.903}{\scriptsize(0.055)} && 
\textbf{0.293}{\scriptsize(0.024)} & 0.307{\scriptsize(0.050)} & \textbf{0.260}{\scriptsize(0.018)} & \textbf{0.416}{\scriptsize(0.080)} \\
\midrule

\multirow{5}{*}{\meltTemp}
& \textsc{Simple} & 
43.0{\scriptsize(2.9)} & 32.6{\scriptsize(0.5)} 
& 32.4{\scriptsize(0.3)} & 61.2{\scriptsize(8.2)} && 
23.5{\scriptsize(1.5)} & 18.9{\scriptsize(0.5)} 
& 15.7{\scriptsize(0.8)} & 40.2{\scriptsize(7.1)} \\
& \textsc{Dropout} & 
40.6{\scriptsize(0.7)} & 32.9{\scriptsize(0.7)} & 31.5{\scriptsize(1.7)} & 55.0{\scriptsize(1.1)}  && 
22.1{\scriptsize(0.5)} & 19.2{\scriptsize(1.0)} & 15.9{\scriptsize(0.9)} & \textbf{33.0}{\scriptsize(1.7)} \\
& \textsc{Certi} &  
40.7{\scriptsize(0.8)} & \textbf{31.6}{\scriptsize(1.5)} & 
\textbf{30.0}{\scriptsize(1.7)} & 57.5{\scriptsize(1.4)} && 
22.0{\scriptsize(1.5)} & 18.9{\scriptsize(1.4)} 
& \textbf{14.6}{\scriptsize(1.7)} & 35.3{\scriptsize(2.2)} \\
& \textsc{DER} & 
70.7{\scriptsize(12.1)} & 36.5{\scriptsize(1.3)} & 60.6{\scriptsize(19.5)} & 110.6{\scriptsize(18.4)} && 
47.3{\scriptsize(10.7)} & 24.6{\scriptsize(1.3)} & 44.5{\scriptsize(21.1)} & 95.0{\scriptsize(20.8)} \\
& \textsc{GRation} & 
\textbf{40.3}{\scriptsize(1.0)} & 32.5{\scriptsize(1.4)} 
& 31.3{\scriptsize(1.1)} & \textbf{54.7}{\scriptsize(1.8)} &
& \textbf{21.7}{\scriptsize(1.1)} & \textbf{18.2}{\scriptsize(1.0)} 
& 15.2{\scriptsize(1.0)} & 33.9{\scriptsize(2.2)} \\
\midrule

\multirow{5}{*}{\density}
& \textsc{Simple} & 
63.9{\scriptsize(6.4)} & 50.6{\scriptsize(4.2)} & 
46.0{\scriptsize(3.0)} & 91.0{\scriptsize(14.6)} && 
34.1{\scriptsize(4.4)} & 28.3{\scriptsize(3.6)} & 
26.5{\scriptsize(2.9)} & 50.9{\scriptsize(11.8)} \\
& \textsc{Dropout} & 
\textbf{55.4}{\scriptsize(1.5)} & 50.2{\scriptsize(2.0)} & 45.3{\scriptsize(2.9)} & \textbf{68.7}{\scriptsize(3.9)} && 
\textbf{28.1}{\scriptsize(3.6)} & \textbf{24.9}{\scriptsize(1.1)} & 24.8{\scriptsize(4.3)} & \textbf{35.3}{\scriptsize(7.1)} \\
\multirow{3}{*}{(scaled:$ \times 1,000$)} 
& \textsc{Certi} &  
56.7{\scriptsize(3.0)} & \textbf{49.8}{\scriptsize(4.6)} & 45.4{\scriptsize(1.1)} & 72.6{\scriptsize(8.0)} && 
28.6{\scriptsize(1.9)} & 26.1{\scriptsize(1.8)} & 24.1{\scriptsize(0.7)} & 36.3{\scriptsize(6.7)} \\
& \textsc{DER} & 
252.4{\scriptsize(85.7)} & 227.2{\scriptsize(104.2)} & 219.6{\scriptsize(81.4)} & 302.9{\scriptsize(74.0)} && 
165.3{\scriptsize(68.8)} & 162.5{\scriptsize(94.8)} & 139.8{\scriptsize(60.4)} & 201.6{\scriptsize(45.4)} \\
& \textsc{GRation} & 
55.9{\scriptsize(4.8)} & 50.4{\scriptsize(10.0)} & \textbf{44.3}{\scriptsize(3.1)} & 70.8{\scriptsize(4.0)} && 
29.1{\scriptsize(3.6)} & 29.4{\scriptsize(9.0)} & \textbf{23.2}{\scriptsize(2.6)} & 35.9{\scriptsize(3.9)} \\
\midrule

\multirow{5}{*}{\oxygen}
& \textsc{Simple} & 
170.2{\scriptsize(6.2)} & 8.7{\scriptsize(8.4)} & 26.5{\scriptsize(28.4)} & \textbf{419.3}{\scriptsize(31.4)} && 
7.2{\scriptsize(0.5)} & 2.3{\scriptsize(0.2)} & 4.8{\scriptsize(1.2)} & \textbf{29.8}{\scriptsize(0.8)}  \\
& \textsc{Dropout} & 
168.7{\scriptsize(7.4)} & 7.5{\scriptsize(4.3)} & 14.1{\scriptsize(3.5)} & 424.3{\scriptsize(20.9)} && 
7.3{\scriptsize(1.6)} & 2.4{\scriptsize(0.8)} & \textbf{4.3}{\scriptsize(1.5)} & 30.4{\scriptsize(3.2)} \\
& \textsc{Certi} &  
181.9{\scriptsize(21.2)} & \textbf{4.9}{\scriptsize(1.6)} & \textbf{12.4}{\scriptsize(1.1)} & 462.5{\scriptsize(56.9)} && 
8.3{\scriptsize(1.5)} & 2.4{\scriptsize(0.6)} & 5.4{\scriptsize(1.7)} & 38.5{\scriptsize(3.6)} \\
& \textsc{DER} & 
247.0{\scriptsize(24.9)} & 26.1{\scriptsize(10.9)} & 24.4{\scriptsize(8.1)} & 604.3{\scriptsize(61.8)} && 
25.0{\scriptsize(8.9)} & 15.5{\scriptsize(8.2)} & 15.3{\scriptsize(6.3)} & 58.6{\scriptsize(7.4)} \\
& \textsc{GRation} & 
\textbf{168.4}{\scriptsize(22.4)} & 7.0{\scriptsize(6.4)} & 14.8{\scriptsize(4.3)} & 423.7{\scriptsize(52.7)} && 
\textbf{7.0}{\scriptsize(1.4)} & \textbf{2.1}{\scriptsize(0.5)} & 4.4{\scriptsize(1.6)} & 31.7{\scriptsize(5.9)} \\

\bottomrule[1.5pt]
\end{tabular}}
\end{table*}
\paragraph{Additional results on the effect of balancing data and label-anchored mixup}
\cref{table:ablation_all_mol} and~\cref{table:ablation_all_plym} present studies on the effect of balancing data and different options in the label-anchored mixup augmentation for molecules and polymers, respectively. They provide more evidence to our observations that (1) the effect of our pseudo-labeling method ($\mathcal{G}_{\text{imb}}\cup\mathcal{G}_{\text{conf}}$) about improving the model performance in the entire label range and the \fewrg; (2) the essential role of the regression confidence $\sigma$ and reverse sampling rate $p$ in our pseudo-labeling about improving pseudo-label quality and reducing imbalance label bias; and (3) the complementary effect of $\mathcal{H}_{\text{aug}}$ about approximating the perfect balance of the training distribution.

\paragraph{Additional results on the regression confidence measurements}
\cref{table:uncertainty_all} show all comparisons among different confidence measurements. \textsc{GRation} consistently performs best in the entire label range excepting dataset \density on which \textsc{Dropout} is slightly better than \textsc{GRation}.

\end{document}